\documentclass{article}
\pdfoutput=1 
\usepackage[utf8]{inputenc}
\usepackage{amsmath}
\usepackage{amstext}
\usepackage{enumerate}
\usepackage[T1]{fontenc}

\usepackage{graphicx}
\usepackage{tikz,pgfplots}
\usepgfplotslibrary{groupplots}
\pgfplotsset{compat=newest}
\usetikzlibrary{shapes.geometric}

\usepackage{listings}
\usepackage{amsmath}
\usepackage{colortbl}
\definecolor{maroon}{cmyk}{0,0.87,0.68,0.32}

\usepackage{multicol}
\usepackage{multirow}
\usepackage{booktabs}

\usepackage{xcolor}
\usepackage{afterpage}
\usepackage{geometry}
\usepackage{changes}
\usepackage[shortlabels]{enumitem}
\usepackage{color}
\usepackage{colortbl}
\usepackage[square,numbers]{natbib}
\usepackage{authblk}
\usepackage{url}

\usepackage{caption}
\usepackage{subcaption}
\usepackage{xspace}
\usepackage{float}
\usepackage{rotating}
\usepackage{appendix}

\newcommand{\virgolette}[1]{{``#1''}}
\newcommand{\Het}{{\em Het-node2vec}\xspace}

\newcommand{\RW}{random walk\xspace}
\newcommand{\so}{{second order}\xspace}
\newcommand{\ntv}{node2vec\xspace}

\usepackage{xparse}

\newlength\WIDTHOFBAR
\NewDocumentCommand{\blackwhitebar}{ O{1} O{0} m }{ % max min value
  \pgfmathsetmacro\x{(#3-#2)/#1}
  #3 {\color{black!100}\rule{\x cm}{8pt}}{\color{black!30}\rule{\WIDTHOFBAR - \x cm}{8pt}}
}

\title{Het-node2vec: second order random walk sampling for heterogeneous multigraphs embedding}

\author[1]{Mauricio Soto-Gomez}
\author[4,5,6]{Peter Robinson}
\author[3]{Carlos Cano}
\author[1]{Ali Pashaeibarough}
\author[1]{Emanuele Cavalleri}
\author[2]{Justin Reese}
\author[1,2]{Marco Mesiti}
\author[1,5]{Giorgio Valentini}
\author[1,2,5]{Elena Casiraghi}
\affil[1]{AnacletoLab, Dipartimento di Informatica, Universit\`a degli Studi di Milano, Italy}
\affil[2]{Lawrence Berkeley National Laboratory, USA}
\affil[3]{Department of Computer Science and Artificial Intelligence, University of Granada, Spain}
\affil[4]{Jackson Lab for Genomic Medicine, Farmington, CT, USA}
\affil[5]{ELLIS - European Laboratory for Learning and Intelligent Systems}
\affil[6]{Berlin Institute of Health - Charité, Universitätsmedizin, Berlin, Germany}
\date{}

\definechangesauthor[name={Peter Robinson}, color=blue]{PR}
\definechangesauthor[name={Giorgio Valentini}, color=blue]{GV}
\definechangesauthor[name={Elena Casiraghi}, color=red]{EC}
\definechangesauthor[name={Elena Casiraghi}, color=purple]{nota-EC}
\definechangesauthor[name={Mauricio Soto}, color=violet]{MS}

\begin{document}

\maketitle

\begin{abstract}
Many real-world problems are naturally modeled as heterogeneous graphs, where nodes and edges represent multiple types of entities and relations.
Existing learning models for heterogeneous graph representation usually depend on the computation of specific and user-defined heterogeneous paths, or in the application of large and often not scalable deep neural network architectures.
We propose \Het, an extension of the \ntv~algorithm, designed for embedding heterogeneous graphs.
\Het addresses the challenge of capturing the topological and structural characteristics of graphs and the semantic information underlying the different types of nodes and edges of heterogeneous graphs, by introducing a simple stochastic node and edge type switching strategy in second order random walk processes. 
The proposed approach also introduces an `attention mechanism' to focus the random walks on specific node and edge types, thus allowing more accurate embeddings and more focused predictions on specific node and edge types of interest.
Empirical results on benchmark datasets show that \Het achieves comparable or superior performance with respect to state-of-the-art methods for heterogeneous graphs in node label and edge prediction tasks.
\end{abstract}

\section{Introduction}

In the field of biology, medicine, social science, economy, and many other disciplines, the representation of relevant problems through complex graphs of interrelated concepts and entities motivates the increasing interest of the scientific community towards Network Representation Learning~\citep{Zhang20}.
Indeed, by learning low-dimensional representations of network vertices that reflect the network topology and the structural  relationships between nodes, 
we can translate the non-Euclidean graph representation of nodes and edges into a fully Euclidean embedding space that can be easily ingested into vector-based machine learning algorithms to efficiently carry out network analytic tasks, ranging from vertex classification and edge prediction to unsupervised clustering, node visualization, and recommendation systems~\citep{Grover16,Martinez17,Wang18,Wang17,Zhang16}.

To this aim, in the past decade most of research efforts focused on homogeneous networks, by proposing matrix factorization-based methods~\citep{Natarajan14}, random walk based methods~\citep{Perozzi14,Grover16}, edge modeling methods~\citep{Jian15}, Generative Adversarial Nets~\citep{Wang19}, and deep learning methods~\citep{Cao16,Hamilton17}.

Nevertheless, the highly informative representation provided by graphs that include different types of entities and relationships motivates the development of increasingly complex networks, also including Knowledge Graphs~\citep{Dai20, cavalleri2024rna}, sometimes referred as multiplex-heterogeneous networks~\citep{valdeolivas2019random}, or simply as heterogeneous networks~\citep{Dong20}, where different types of nodes and edges are used to integrate and represent the information carried by multiple sources of information. Following these advancements, Heterogeneous Graph Representation Learning (HGRL) algorithms have been recently proposed to process such complex, heterogeneous graphs~\citep{yang2020heterogeneous,Xie21,bing2023heterogeneous}.

The core issue with HGRL is to simultaneously capture  the structural properties of the network and the semantic properties of the heterogeneous nodes and edges; in other words, we need node and edge type-aware embeddings 
that can preserve both the structural and the semantic properties of the underlying heterogeneous graph.

In this context, from an algorithmic point of view, three main lines of research have recently emerged, both inspired by homogeneous network representation learning~\citep{Dong20}: the first one leverages results obtained by homogeneous \textit{Random-Walk (RW) based approaches}; they are based on the \virgolette{distributional hypothesis}\footnote{The distributional hypothesis was originally proposed in linguistics \citep{fries1954meaning,harris1954distributional}. It assumes that \virgolette{linguistic items with similar distributions have similar meanings}, from which it follows that words (elements) used and occurring in the same contexts tend to purport similar meanings~\citep{harris1954distributional}.}, firstly exploited to capture the semantic similarity of words~\citep{Mikolov13}, and then extended to capture the similarity between graph nodes~\citep{Grover16}; the second one exploits neural networks specifically designed to process graphs, using e.g., convolutional filters~\citep{Kipf17}, and more generally direct supervised feature learning through \textit{Graph Neural Networks (GNNs)}~\citep{Hamilton17}.
The third research line \cite{bordes2011learning,Yang2014EmbeddingEA,wang2014knowledge,trouillon2016complex} views heterogeneous graphs as knowledge bases expressing relationships (edges) between sources and destination nodes, and then learns the latent space where all such relationships are ``optimally'' represented.

\textit{RW-based methods} share the assumption that nodes having the same structural context or being topologically close in the network (homophily) are also close in the embedding space.
Some of these methods separately process each homogeneous network included in the original heterogeneous graph. 

As an example, in~\citep{Tang15} the heterogeneous network is first projected into several homogeneous bipartite networks; then, an embedding representing the integrated multi-source information is computed by a joint optimization technique combining the skip-gram models individually defined on each homogeneous graph.
A similar decomposition is initially applied in \citep{Zitnik17}, where the original heterogeneous graph is split into a set of hierarchically structured homogeneous graphs. Each homogeneous graph is then processed through node2vec~\citep{Grover16}, and the embedding of the heterogeneous network is finally obtained by using recursive regularization, which encourages the different embeddings to be similar to their parent embedding.
Another approach in this context constraints the RWs used to collect node contexts for the embeddings into specific {\em meta-paths}: the walker can step only between pre-specified pairs of vertices, thus better capturing the structural and semantic characteristics of the nodes~\citep{Dong17}. Other related approaches combine vertex pair embedding with meta-path embeddings~\citep{Park19}, or improves the heterogeneous Spacey RW algorithm by imposing meta-paths, graphs and schema constraints~\citep{He19}.

Differently from the distributional hypothesis approach that usually applies shallow neural networks to learn the embeddings, \textit{GNN approaches} apply deep neural-network encoders to provide more complex representations of the underlying graph~\citep{Wu20}. 
By this approach, the deep neural network recursively aggregates information from neighborhoods of each node in such a way that the node neighborhood itself defines a computation graph that learns how to propagate information across the graph to compute the node features~\citep{Hamilton17,Gilmer17}. As it often happens for the distributional approach, the usual strategy used by GNNs to deal with heterogeneous graphs is to decompose them into its homogeneous components. For instance, Relational Graph Convolutional Networks~\citep{Schlichtkrull18}  maintain distinct weight matrices for each different edge type,
or Heterogeneous Graph Neural Networks~\citep{Zhang19a}, apply first-level Recurrent Neural Networks (RNN) to separately encode features for each type of neighbour nodes, and then a second level RNN to combine them. 
Also Decagon~\citep{Zitnik18}, which has been successfully applied to model polypharmacy side effects, uses a graph decomposition approach by which node embeddings are separately generated by edge type and the resulting computation graphs are then aggregated.
Other approaches add meta-path edges to augment the graph~\citep{Wang19bHAN} or learn attention coefficients that weight the importance of different types of vertices~\citep{Chen18}. The drawback of all the aforementioned GNN approaches is that some relations may not have sufficient occurrences, thus leading to poor relation-specific weights in the resulting GNN. To overcome this problem, an Heterogeneous GNN~\citep{Hu20} that uses the Transformer-like self-attention architecture have been recently proposed.

\textit{Relation-learning approaches} \citep{bordes2011learning,Yang2014EmbeddingEA,wang2014knowledge,trouillon2016complex} use contrastive learning techniques to project entities (head and tail nodes) and the relationships (edges) between them into low-dimensional latent spaces that preserve the relationships between entities and relationships. This is achieved by assigning a score to each (head, relation, tail) triple, which is maximized for true triples and minimized for ``corrupted triples'', that is triples not truly existing in the graph.

The simplest yet effective relation-learning technique is DistMult \cite{Yang2014EmbeddingEA}; it projects triples into a latent space where the score maximized for true triples is the generalized dot product between the embeddings of the source node, destination node, and the edge connecting them. 
ComplEx~\cite{trouillon2016complex} is an extension of DistMult that can deal with oriented relationships. To achieve this, it projects the graph entities into a complex space where the score maximized for true triples is computed as the generalized dot product in the complex space. 
TransE~\citep{bordes2011learning} and its extension to hyperplanes, TransH~\citep{wang2014knowledge}, are probably the most popular relation-learning techniques. TransE, projects triples into a latent space where, for true triples, the relation edge between two nodes is modeled as a translation vector between the source vector and the destination. 
The definition of TransE collapses all the triples into a unique latent space. This might decrease separability between relationships; therefore, TransH extends TransE by finding one hyperplane for each relation type.

Despite the impressive advancements achieved in recent years by the aforementioned methods (distributional approaches, GNN-based, and relation learning approaches), they show drawbacks and limitations.

Indeed, methods based on the distributional hypothesis, which base the embeddings on the random neighborhood sampling, usually rely on the manual exploration of heterogeneous structures, i.e., they require human-designed meta-paths to capture the structural and semantic dependencies between the nodes and edges of the graph. 
This requires human intervention and non-automatic pre-processing steps for designing the meta-paths and the overall network scheme.
Moreover, similarly to Heterogeneous GNN, in most cases, they treat separately each type of homogeneous network extracted from the original heterogeneous one and are not able to focus on specific types of nodes or edges that constitute the objective of the underlying prediction task (e.g., prediction of a specific edge type).

For what regards GNN,  an open issue is represented by their computational complexity, which is exacerbated by the intrinsic complexity of heterogeneous graphs, thus posing severe scaling limitations when dealing with big heterogeneous graphs.
Moreover, in most cases, heterogeneous GNN models use different weight matrices for each type of edge or node, thus augmenting the complexity of the learning model.  Some GNN methods augment the graphs by leveraging human-designed meta paths, thus showing the same limitation of distributional approaches, i.e., the need for human intervention and non-automatic pre-processing steps. 

Relation-learning approaches handle each triple with equal probability, leading to a bias towards the most represented relationships. Furthermore, while models like TransH attempt to induce better separability between relationships, they struggle with 1-to-N or N-to-N relationships, where two node types are linked by semantically different relationships \cite{wang2021kg2vec}. More complex triple-learning techniques, such as ComplEx~\cite{trouillon2016complex}, suffer from computational inefficiency, hindering their practical application.

To overcome some of these drawbacks, we propose a general framework to deal with complex heterogeneous networks, in the context of the  previously discussed "distributional hypothesis" random-walk based research line.
The proposed approach, which we named {\em Het-node2vec} to remark its derivation from the classical {\em node2vec} algorithm~\citep{Grover16}, can process heterogeneous multi-graphs characterized by multiple types of nodes and edges and can scale up with big networks, due to its intrinsic parallel nature. 
{\em Het-node2vec} does not require manual exploration of heterogeneous structures and meta-paths to deal with heterogeneous graphs but directly models the heterogeneous graph as a whole without splitting the heterogeneous graph into its homogeneous components.
It can focus on specific edges or nodes of the heterogeneous graph, thus introducing a sort of ``attention'' mechanism~\citep{Bahdanau15}, conceptually borrowed from the deep neural network literature, but performed in an original and simple way in the world of RW visits of heterogeneous graphs.
Our proposed approach is particularly appropriate when we need to predict edge or node types that are underrepresented in the heterogeneous network, since the algorithm can focus on specific types of edges or nodes, even when they are largely outnumbered by the other types.
At the same time, the proposed algorithms learn embeddings that are aware of the different types of nodes and edges of the heterogeneous network and of the topology of the overall network.

The main contributions of our work are the following:

\begin{itemize}
\item We introduce \Het that extends the \ntv~\cite{Grover16} algorithm to handle heterogeneous graphs. \Het enables the generation of node embeddings that capture both the graph structural topology and the semantic diversity of node and edge types.

\item We propose a type-aware random-walk sampling schema that incorporates node and edge type information, allowing the algorithm to focus on specific types of nodes or edges in heterogeneous graphs; by exploiting the graph semantic properties, our strategy improves the quality of the computed embeddings.
  
\item We design special node-type and edge-type switching mechanisms that can prioritize transitions toward specific node or edge types, allowing the algorithm to address problems where specific node/edge types are crucial for the prediction task under investigation.

\item We propose a computationally efficient implementation of \Het that retains the scalability of the original \ntv algorithm.

\item We conduct experiments on benchmark heterogeneous graphs to show that \Het can improve performance in node classification and link prediction tasks compared to existing methods.

\item To ensure a F.A.I.R. (Findable, Accessible, Interoperable, and Reusable) description and encourage further research and applications involving heterogeneous graphs, we release an open-source implementation of \Het, and we integrate it within the efficient GRAPE library~\cite{grape23}.
\end{itemize}

%%%%%%%%%%%%%%%%%%%%%%%%%%%%%%%%%%%%%%%%%
\section{Homogeneous RW based methods}

DeepWalk~\cite{Perozzi14} and \ntv~\cite{Grover16} algorithms construct node and edge embeddings that represent the topology of a graph by leveraging the word2vec paradigm~\cite{Mikolov13}. They ``linearize'' the graph via first or second order RWs across the graph; while DeepWalk exploits first order RWs, \ntv uses second order RWs to capture either the structural or the homophilic similarities in the graph \cite{Grover16}. The encoded pairs of nodes generated trough RW samples then feed a shallow neural network-based embedding algorithm, such as Skipgram or CBOW, to obtain vectorial representations of the graph nodes.

More precisely, let $X_t$ represent the node visited at step $t$ in a RW on graph $G=(V,E)$, where $V$ and $E$ denote, respectively, the set of nodes and the set of edges.
In DeepWalk, at step $t$, the transition from node $X_t=v$ to one of its neighbors $X_{t+1}=x \in \mathcal{N}(v)$, being $\mathcal{N}(v)$ the set of one-hop neighbors of $v$, is governed by a probability that only accounts for weight $w_{vx}$, over the edge $vx$ that connects $v$ and $x$, that is $P(X_{t+1}=x | X_t=v) \; \propto\; \pi_{vx}=w_{vx}$.

\begin{figure}[bth]
\centering
\includegraphics[width=.4\linewidth]{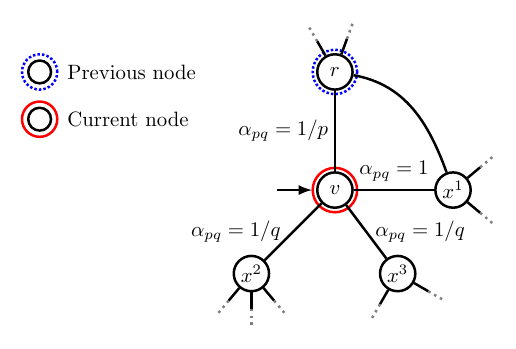}
\caption{A step (at time $t$) of a \so RW in a homogeneous graph. At time $t-1$ the \RW was in $X_{t-1}= r$, and has just moved from node $r$ to node $v$. Then, the probability of moving from $v$ to any nearest-neighbor is proportional to $\alpha_{pq}\cdot w_{vx}$, where $w_{vx}$ denotes the weight of edge $vx$, and $\alpha_{pq}$ depends on a return parameter $p$ and on an outward parameter $q$.}
\label{fig:node2vec}
%\Description{node2vec schema}
\end{figure}

Node2vec extends DeepWalk by biasing the walk via a second order transition probability $P(X_{t+1}=x | X_t=v, X_{t-1}=r)$ that depends also on the node visited by the RW in the previous step,  $X_{t-1}=r$. 
In more detail (see also Figure \ref{fig:node2vec}): 

\begin{align}\label{eq:node2vec_prob}
 \pi_{rvx} &= \alpha_{pq}(r,v,x)\cdot w_{vx}\\
   \alpha_{p,q} (r,v,x) &=
\begin{cases} 
\frac{1}{p} & \text{ if } d_{rx}=0 \\
1           & \text{ if } d_{rx}=1 \\
\frac{1}{q} & \text{ if } d_{rx}=2
\end{cases} \label{eq:alpha}\\
 P(X_{t+1}=x | X_t=v, X_{t-1}=r) &= \pi_{rvx}/{\textstyle \sum\limits_{vz\in E}\pi_{rvz}}   \label{eq:TransitionProb}     
\end{align}

where $\pi_{rvx}$ denotes the unnormalized transition probability and  $\alpha_{pq}$ is a parametric function depending on the hop-distance $d_{rx}$ between nodes $r$ and $x$.
 
Hyperparameters $p$ (\textit{return} or \textit{inward} hyperparameter) and $q$ (\textit{in-out} or \textit{explore}) bias the walk to either favor a depth-first (DFS)-like or breadth-first (BFS)-like search, thereby controlling the tendency of the RW to explore new regions or revisit nodes of the graph, respectively. The transition probability is finally obtained as the normalization of the function  $\pi_{rvx}$ with respect to the set of neighbors of $v$ (Equation \ref{eq:TransitionProb}).
 These dynamics are illustrated in Figure~\ref{fig:node2vec}, which shows a step of a second order RW. Supplementary section~\ref{app:pq_effects} 
 reports some experimental examples showing the effect of parameters $p$ and $q$ in the embedding on graphs characterized by simple topological structures.

%%%%%%%%%%%%%%%%%%%%%%%%%%%%%%%%%%%%%%%%%%%%%%%%%%%%%%%%%%%%%%% 
\section{\Het: Heterogeneous node2vec}
\label{sec:HetGraph}

In real-world scenarios, complex information is often structured as a network of complex and diverse relationships between entities (or concepts) that belong to multiple distinct classes. Modeling these relationships as heterogeneous graphs provides a natural framework for representing such interconnected data. By capturing the multiplicity of entity types and relationships, heterogeneous graphs enable the automated extraction of knowledge and insights from the underlying data.

\begin{figure}[htbp]
\centering
\includegraphics[width=.6\linewidth]{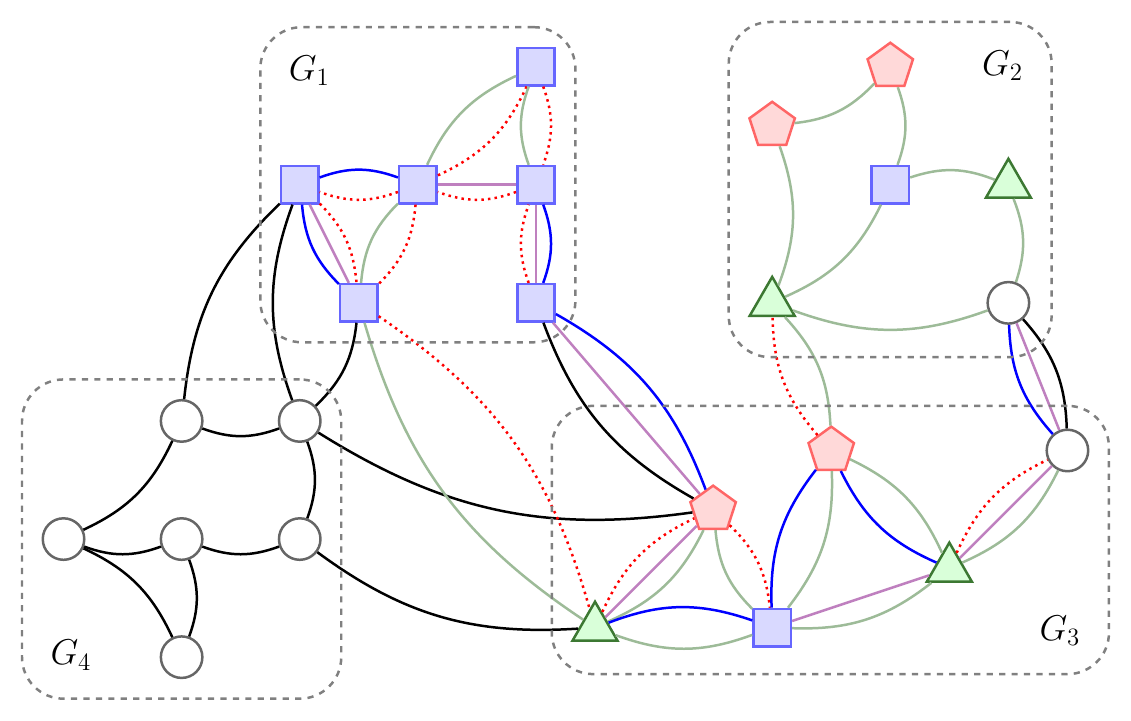}
\caption{A heterogeneous multigraph with nodes and edges of different types. Different colors are used to represent node and edge types. Multiple types of edges may connect the same pair of nodes. 
}\label{fig:HetNet}
%\Description{Example of different types of heterogeneous networks}
\end{figure}

Figure~\ref{fig:HetNet} represents a heterogeneous network with different types of nodes and edges.
Different colors, shapes, and strokes represent different types of nodes and edges.
The graph in the figure contains four different subgraphs depicting some classes of graphs that can be processed with \Het.
Subgraph $G_1$ is a multigraph having nodes of the same type, but edges of different types (edges having different colors), and multiple edges may connect the same pair of nodes. Subgraph $G_2$ is a graph having different types of nodes, but the same type of edges. Subgraph $G_3$ is a multigraph having both different types of nodes and edges, and the same pair of nodes may be connected by multiple edges.
Finally, subgraph $G_4$ is a graph with both homogeneous nodes and edges: for this subgraph, the classical {\em node2vec} suffices since this algorithm can be applied to graphs having homogeneous nodes and edges.

In the following, we will show how \Het is able to manage both
heterogeneous networks and multigraphs, i.e. graphs where the same pair of nodes may be connected by multiple types of edges.

%%%%%%%%%%%%%%%%%%%%%%%%%%%%%%%%%%%%%%%

\subsection{The basic \Het~algorithm}
\Het generalizes the \ntv algorithm by sampling type-aware RWs, and adopting a sort of  ``attention'' mechanism~\citep{Velickovic2018} allowing to generate type-aware RW samples.
This can be accomplished by introducing ``switching'' parameters that control the way the RW moves between different node and edge types, providing a mechanism to incorporate the semantic information of the graph into the RW generation process.

Let $G=(V,E)$ be a heterogeneous multigraph where $\phi: V \rightarrow \Sigma_\phi$ denotes the function defining the type of a node, and $\psi:E \rightarrow \Sigma_\psi$ denotes the function defining the type of an edge.
Let $E_{t+1}$ denote the edge traversed by the RW from the node  $X_{t}$ to node $X_{t+1}$.
Consider a RW currently residing at node $X_t = v$, coming from node $X_{t-1} = r$  through the edge $E_{t} =e_{rv}$;  if $x$ is a neighbor of $v$, \Het defines the \so  transition probability of stepping to $X_{t+1} = x$ through an edge $E_{t+1} = e_{vx}$ as:

\begin{equation}
  P\Bigl(X_{t+1}=x, E_{t+1} = e_{vx} | X_t=v, X_{t-1}=r, E_{t} = e_{rv} \Bigr)\; \propto\; \hat\pi_{rvx,e_{rv}e_{vx}}=\Phi_{sc}\cdot \alpha_{pq}\cdot  w_{e_{vx}}, \label{eq:hetnode2vec_prob_multi}
\end{equation}

where $\hat\pi_{rvx,e_{rv}e_{vx}}$ denotes the unnormalized transition probability, function $\alpha_{pq}$ is defined as in \ntv, and $w_{e_{vx}}$ is the weight over the edge $e_{vx}$ connecting $v$ and $x$. 
With respect to \ntv (Equation~\ref{eq:node2vec_prob}), the transition probability in Equation~\ref{eq:hetnode2vec_prob_multi} incorporates a new term, $\Phi_{sc}$, which depends on both the type of the nodes and the type of the edges involved in the transition and is defined as:

$$\Phi_{sc} = \beta_s(v,x) \cdot \gamma_c(e_{rv},e_{vx}),$$

\begin{figure}[htb]
  \centering
     \begin{subfigure}[b]{.67\linewidth}
     \centering
       \includegraphics[width=.43\linewidth]{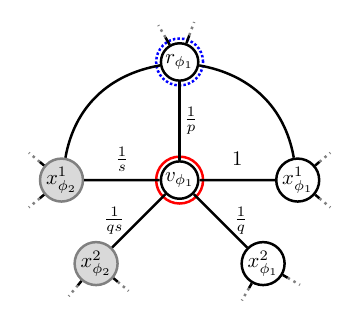}
     \hspace{10pt}
     \includegraphics[width=0.43\linewidth]{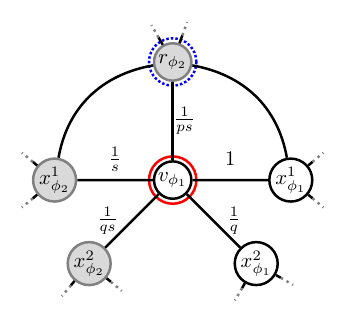}
     \caption{}%\HeHontv.}
     \label{fig:heho_transition_prob}
   \end{subfigure}%
   ~
   \begin{subfigure}[b]{.3\linewidth}
     \centering
     \includegraphics[width=.9\linewidth]{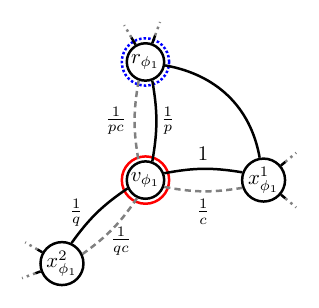}
     \caption{}%\HoHentv.}
     \label{fig:hohe_transition_prob}
 \end{subfigure}
    
 \begin{subfigure}[b]{\linewidth}
     \centering
     \hfill \includegraphics[width=.25\linewidth]{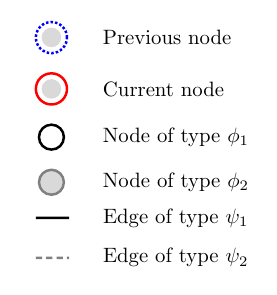}
     \hspace{-10pt}
     \hfill \includegraphics[width=.33\linewidth]{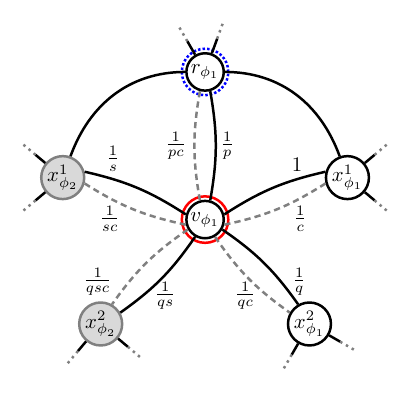}
     %\hspace{8pt}
     \includegraphics[width=0.33\linewidth]{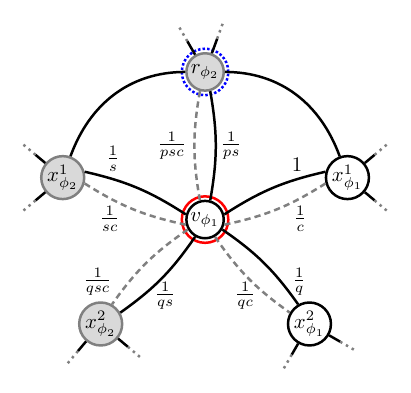}
     \caption{}%\HeHentv.}
     \label{fig:hehe_transition_prob}
   \end{subfigure}
   \caption{
   Unnormalized transition probabilities for \Het~in unweighted heterogeneous networks with:  (a) heterogeneous nodes and homogeneous edges, i.e. $c=1$; (b) heterogeneous edges and homogeneous nodes, i.e. $s=1$; (c) both nodes and edges are heterogeneous.
     The color of nodes and the line-style (dashed or continuous) of edges represent their type. 
      To simplify the notation the edges are unweighted and nodes $x$ with type $\phi_*$ are denoted as $x_{\phi_*}$.
     Edge labels indicate the value of the function $\Phi_{sc}\cdot \alpha_{pq}$ (without considering edge weights) for a second order RW starting from $v_{\phi_1}$, i.e. $X_t = v_{\phi_1}$ and coming from node $r$, i.e. $X_{t-1} = r_{\phi_1}$ or $X_{t-1} = r_{\phi_2}$. 
     }
  \label{fig:het_transition_prob}
%\Description{Un-normalized Het-node2vec transition probabilities on different types of heterogeneous networks}
\end{figure}

The parametric functions $\beta_s$ and $\gamma_c$ bias the RW when switching the node-type and edge-type, respectively (Figure~\ref{fig:het_transition_prob}). 
Notice that the function $\beta_s$ depends on the types of nodes  $v$ and $x$, while $\gamma_c$ depends on the types of edges $e_{rv}$ and $e_{vx}$\footnote{For the sake of simplicity, we will omit the arguments of functions $\Phi_{sc}$, $\beta_s$ and $\gamma_c$ whenever the context is clear.}. 
More precisely, we set $\beta_s$ and $\gamma_c$ to increase or decrease the probability of changing the node/edge types in the next step of the RW as follows:
\begin{equation} 
  \label{eq:beta_gamma_generic}
       \beta_s(v,x)=   
        \begin{cases}
          \frac{1}{s} & \text{ if } \phi(v) \neq \phi(x) \\
          1           & \text{ otherwise } 
        \end{cases}
\qquad \text {and} \qquad 
       \gamma_c(e_{rv},e_{vx})=   
        \begin{cases}
          \frac{1}{c} & \text{ if } \psi(e_{rv}) \neq \psi(e_{vx}) \\
          1           & \text{ otherwise } 
        \end{cases}        
\end{equation}
where $s$ (\textit{node switching} weight) and $c$ (\textit{edge switching} weight) are user-set parameters whose values allow biasing the transition probabilities so that the RW explores the graph by preserving ($s>1$, $c>1$) or switching ($s<1$, $c<1$) the types of nodes and edges. 

Summarizing, according to Equation~\ref{eq:hetnode2vec_prob_multi}, and omitting the weights to simplify the notation, we can obtain the following scheme for the \so RW type-aware computation of $\Phi_{sc}\cdot \alpha_{pq}$:

 \begin{equation}\label{eq:HeHe}
   \Phi_{sc}\cdot \alpha_{pq} =
   \begin{cases}
     \text{if }  \psi(e_{rv}) = \psi(e_{vx}): &
     \begin{cases}
     \text{if  }\phi(x) = \phi(v): & 
        \begin{cases}
          \frac{1}{p}  & \text{ if }\: d_{r,x}=0\\[2pt] 
          1            & \text{ if }\: d_{r,x}=1\\[2pt] 
          \frac{1}{q}  &\text{ if }\: d_{r,x}=2
        \end{cases}\\ \\[-5pt]
        \text{otherwise}: &
        \begin{cases}
          \frac{1}{ps} & \text{ if }\: d_{r,x}=0\\[2pt]
          \frac{1}{s}  & \text{ if }\: d_{r,x}=1\\[2pt]
          \frac{1}{qs} & \text{ if }\: d_{r,x}=2          
        \end{cases}
      \end{cases}\\ \\[-5pt]    
           \text{otherwise}:&  
     \begin{cases} \\
     \text{ if  }\phi(x) = \phi(v): & 
        \begin{cases}
           \frac{1}{pc}  & \text{ if }\: d_{r,x}=0\\[2pt] 
           \frac{1}{c}   & \text{ if }\: d_{r,x}=1\\[2pt] 
           \frac{1}{qc}  &\text{ if }\: d_{r,x}=2
        \end{cases}\\ \\[-5pt]
        \text{otherwise}: &
        \begin{cases}
             \frac{1}{psc} & \text{ if }\: d_{r,x}=0\\[2pt]
             \frac{1}{sc}  & \text{ if }\: d_{r,x}=1\\[2pt]
             \frac{1}{qsc} & \text{ if }\: d_{r,x}=2    
        \end{cases}
      \end{cases}
    \end{cases}  
\end{equation}

%\smallskip

For the final computation of the unnormalized transition probabilities, we need to multiply $\Phi_{sc}\cdot \alpha_{pq}$ by $w_{e_{vx}}$(Equations \ref{eq:hetnode2vec_prob_multi} and \ref{eq:HeHe}).
Figure~\ref{fig:het_transition_prob} depicts the unnormalized transition probabilities computed by an iteration of a \Het~RW~generation process.

%%%%%%%%%%%%%%%%%%%%%%%%%%%%%%%%%%%%%%%%%

Focusing on the most general case of Figure~\ref{fig:hehe_transition_prob}, we note that the previous step from $X_{t-1} = r$ to $X_{t} = v$, may have been performed by using an edge either of type $\psi_1$ or of type $\psi_2$.
At step $t+1$ the walk may either move back to $r$ or away to $x$, by using an edge with the same type of the edge used to move from $r$ to $v$. At this point, if $X_{t+1}=r$, then the \ntv~\so~transition function would be equal to $\alpha_{pq}=1/p$, while \Het parameters $\beta_s$ and $\gamma_c$ depend, respectively, on the node-type of $v$ and $r$, and on the edge-type of the edge $(r,v)$, used to move from $r$ to $v$, and the edge $(v,r)$ used to move back from $v$ to $r$.  
If instead $X_{t+1}=x$, eight cases are possible, depending on: (1) the topological distance $d_{r,x}$ between $r$ and $x$; (2) the node-type of nodes $v_{\phi_1}$ and $x$ or $r$; (3) the edge-type of $(v,x)$.

Note that if we set $s<1$ we promote the switching between different node types, and the opposite is true if $s>1$.  To promote edge switching, the same effect can be obtained with $c<1$.

Our description of the \Het allows implementing a general heterogeneous \so RW embedding technique working on fully heterogeneous multigraphs, i.e. graphs with different types of nodes and edges. 

\subsection{Generic, Multiple, and Special node switching.}
\label{subsub:generic-multiple-special-switch}
The basic \Het~algorithm leverages the parameter $s$ to switch between nodes having different type. We name this node-type switching modality as \textbf{generic switching}.
With this strategy, in a transition from a node $v$ toward a node $x$, the switching probability only depends on $\phi(x)$ being different from $\phi(v)$; there is no focus on the specific node-types being switched.
A similar strategy can also be applied to edge type switching, where the switching probability does not depend on the specific edge types $\psi(e_{rv})$ and $\psi(e_{vx})$, but only on the \virgolette{generic switching} between different edge types, i.e. on the condition $\psi(r,v) \neq \psi(v,x)$.

To induce a specific node or edge switching schema, we can define for each couple of nodes types $(\phi,\phi')\in \Sigma_\phi\times\Sigma_\phi$ a node-switching parameter $s_{\phi \phi'}$; and for each couple $(\psi,\psi')\in  \Sigma_\psi\times\Sigma_\psi$ an edge-switching parameter $c_{\psi\psi'}$, such that
$$ 
\beta_s(v,x) = 1/s_{\phi(v)\phi(x)} 
  \quad \text{and} \quad 
\gamma_c(e_{rv},e_{vx}) = 1/c_{\psi(e_{rv})\psi(e_{vx})}.
$$

This approach, which we name \textbf{multiple switching}, allows controlling the specificity of the {\em switching} process and of the resulting RW by emulating a \virgolette{probabilistic metapath}. 
In fact, it is possible to define a set of switching parameters driving the RW generation process to stochastically follow a predefined sequence of node/edge types.
However, the resulting model introduces an increased hyper-parameter complexity.
Indeed, if the number of nodes and edges types in $\Sigma_{\phi}$ and $\Sigma_{\psi}$ are $N$ and $M$ respectively, the model could require up to $2\left(\binom{N}{2}+\binom{M}{2}\right)=\mathcal{O}(N^2+M^2)$  different switching hyperparameters.

Besides, many real-world applications need to focus on a specific node type or on a subset of specific node types.
Therefore, in this case, the switching strategy should promote or demote transitions toward these nodes.
To achieve this, a \textbf{special node-type switching} strategy may be implemented as a specific case of the multiple switching strategy.

In the special node-type switching strategy, the set of node types is partitioned into two subsets: special node types and non-special node types that we denote by $\Sigma_{\phi_\mathcal{S}}$ and $\Sigma_{\phi_\mathcal{NS}}$ respectively.
Following this partition, we say that a node $x$ is ``special'' if $\phi(x)\in \Sigma_{\phi_\mathcal{S}}$; otherwise, we say that the node $x$ is ``non-special''.
The special node-type switching strategy defines the switching parameters between these two node-type sets to promote or demote the probability of switching towards special nodes.

We propose two flavours of the {\em special-node switching} strategy. The first one uses $\beta_s^{(1)}= \frac{1}{s}$ for any RW step toward a special node, independently from the source node type, thus enforcing walks among special nodes. The second one introduces  $\beta_s^{(2)}= \frac{1}{s}$ only when switching from a non special to a special node, but without biasing RWs between special nodes, thus encouraging also the visit of non special nodes.

\begin{figure}[htbp]
  \centering
  \includegraphics[width=.7\linewidth]{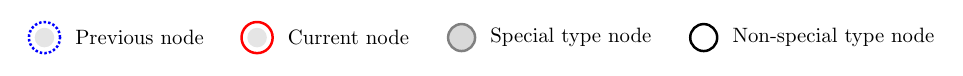}

  \begin{subfigure}[c]{.7\linewidth}
    \centering
    \includegraphics[width=1.07\linewidth]{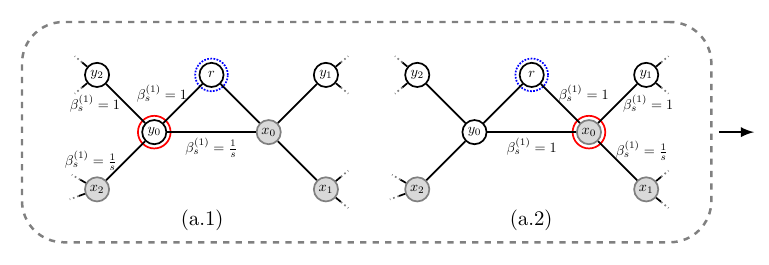}
    \caption{Definition of  $\beta^{(1)}$}
    \label{fig:special_switching1}
    \end{subfigure}
    \begin{subfigure}[c]{0.29\linewidth}
    \centering
    \includegraphics[width=1.2\linewidth]{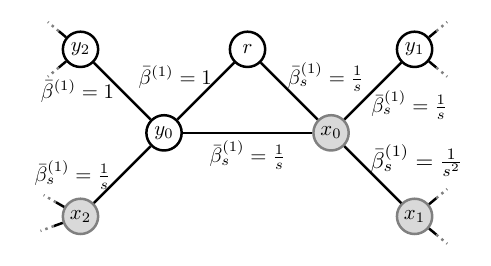}
    \caption{Implementation of  $\beta^{(1)}$}
     \label{fig:special_switching_implementation1}
   \end{subfigure}

  \begin{subfigure}[c]{.7\linewidth}
    \centering
    \includegraphics[width=1.07\linewidth]{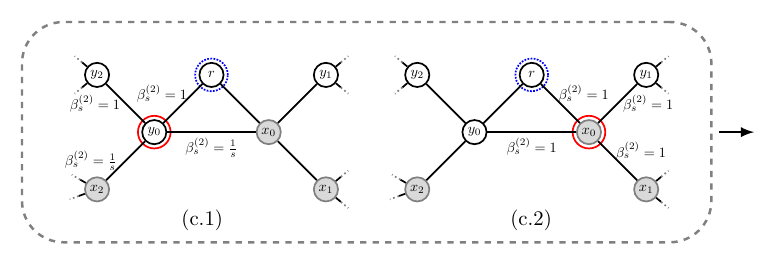}
   \caption{Definition of  $\beta^{(2)}$}
    \label{fig:special_switching2}
    \end{subfigure}
    \begin{subfigure}[c]{0.29\linewidth}
    \centering
    \includegraphics[width=1.2\linewidth]{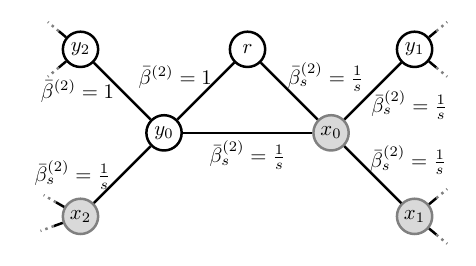}
    \vspace{0pt}
    \caption{Implementation of  $\beta^{(2)}$}
     \label{fig:special_switching_implementation2}
   \end{subfigure}
   \caption{Definition and implementation of special node-type switching strategies.
     (\subref{fig:special_switching1}.1, \subref{fig:special_switching2}.1): When the RW starts from a non-special node, the two strategies bias the transition probability in the same way: both promote/demote transitions towards special node types.
     (\subref{fig:special_switching1}.2): When starting from a special node, the first special node-type switching strategy promotes/demotes the transition toward another special node, independently of the type of the preceding node in the RW.
     (\subref{fig:special_switching2}.2): When starting from a special node type, the second special node-type switching strategy does not bias the choice of the next node; in this way, the RW gains a broader knowledge about the types of its neighborhoods, including also non special nodes.
     (\subref{fig:special_switching_implementation1}, \subref{fig:special_switching_implementation2}): Each node-type switching function is implemented so that it only depends on the edge used in the transition and not on the direction of the RW. 
     In the case of Figure \subref{fig:special_switching_implementation1}, the bias applied to the edge connecting \( x_0 \) and \( x_1 \) is \( 1/s^2 \), which is \( 1/s \) times lower than the bias applied to all other outbound edges, which are set to \( 1/s \). After normalizing these biases to ensure they sum to 1, the probability of switching from \( x_0 \) to \( x_1 \) will be \( 1/s \) times larger (if \( s < 1 \)) or lower (if \( s > 1 \)) than the probability of switching to any other neighbor, according to the \Het design.
 In the case of 1, the bias applied to all the outgoing edges (from \( x_0 \) to each of the neighbors) is constant so that the switching probability resulting after normalizing the switching weights will not promote or demote any edge.
   }
  \label{fig:special_switching} 
%\Description{Un-normalized transition probabilities for the two special type-type switching strategies and their corresponding implementations.}
\end{figure}

Figure~\ref{fig:special_switching} depicts these two special switching strategies.
In the first node-type switching strategy (Equation~\ref{eq:special_node_beta1}, Figure~\ref{fig:special_switching}.a), the node-switching function $\beta^{(1)}_s$ is set to $1/s$ when transitioning towards a special node; otherwise, it is set to one:
\begin{equation}
      \beta^{(1)}_s(v,x) =
      \begin{cases}
        \frac{1}{s} & \text{ if } \phi(x) \in \Sigma_{\phi_\mathcal{S}}\\
        1 & \text{otherwise}.
      \end{cases}%\hspace{0pt}
      \label{eq:special_node_beta1}
    \end{equation}

When the value of $s \ll 1$ (or $s \gg 1$), this strategy may generate RWs that visit mainly or exclusively special nodes (or neglect special nodes), therefore failing to capture the type heterogeneity information that characterizes node neighborhoods.
On the contrary, the second special node-type switching strategy allows to explore a broader context, since the function $\beta^{(2)}_s$  takes the value $1/s$ only when transitioning from a non-special node towards a special node (Equation~\ref{eq:special_node_beta2}, Figure~\ref{fig:special_switching}.b); when, instead, the RW starts from a special node and must decide where to move next, $\beta^{(2)}_s$ is set to 1 to avoid biasing the RW probability towards other special nodes: 
    \begin{equation}
      \beta^{(2)}_s(v,x) =
      \begin{cases}
        \frac{1}{s} & \text{ if } \phi(v) \notin \Sigma_{\phi_\mathcal{S}} \wedge \phi(x) \in \Sigma_{\phi_\mathcal{S}}\\
        1 & \text{otherwise}.
      \end{cases}
      \label{eq:special_node_beta2}
    \end{equation}
  In practice, both strategies handle in the same way the situation when the RW resides at a non-special node and must decide where to move next; they use parameter $s$ to promote/demote moving towards special node-types. When, instead, the RW resides at a special node, the first strategy keeps promoting/demoting switches to other special nodes; the second strategy avoids any bias to allow a heterogeneous walk.  
Empirical results reported in section~\ref{sec:special_node_switching_comparison} of the Supplementary Information show that, indeed, the second option is the most promising, and thus, it is the one we used for all the experiments reported in Section~\ref{sec:exp-res}.

Analogously to the case of special nodes-types, it is possible to define a \textbf{special edge-type switching strategy} that promotes/demotes transitioning through a subset of special edge types $\Sigma_{\psi_{\mathcal{S}}}\subset \Sigma_{\psi}$:
\begin{equation}
       \gamma_c(e_{rv},e_{vx}) = 
       \begin{cases}
           \frac{1}{c} & \text{ if } \psi(e_{vx}) \in \Sigma_{\psi_{\mathcal{S}}}\\
           1 & \text{otherwise.}
       \end{cases}
  \label{eq:special_edge_gamma}
  \end{equation}%

%%%%%%%%%%%%%%%%%%%%%%%%%%%%%%%%%%%%%%%%%%%%%%%

%%%%%%%%%%%%%%%%%%%%%%%%%%%%%%%%%%%%%
\subsection{Implementation of \Het}
\label{sec:implementation}

RW-based embedding algorithms may be efficiently implemented by developing optimized and parallel implementations, which make use of succinct data structures \cite{grape23} and reuse samples across different source nodes, as proposed by \ntv authors themselves \cite{Grover16}.

\Het can be implemented in an optimized manner by redefining, where possible, the type switching function so that it does not depend on the direction of the transition but only depends on the types of nodes $x$ and $v$ at the vertices of each edge, or on the type of the edge used to transition from $v$ to $x$, thus allowing the pre-computation of the functions $\beta_s$ and $\gamma_c$.

More precisely, for the \textbf{generic node-type switching strategy}, the bias $\beta_s$ can be precomputed by multiplying the weights of each edge by a factor $\frac{1}{s}$ if its vertices have different types.
  The edge-type switching parameter $\gamma_c$ also depends on the type of the edge used for the previous step, i.e. for transitioning from $X_{t-1}=r$ to $X_{t}=v$; it can be precomputed by considering the types of edges in the 2-hop neighborhoods of each node.

For what regards the \textbf{special-node type switching strategies}, functions $\beta^{(1)}_s$ and $\beta^{(2)}_s$ are defined based on both the edge used for transitioning as well as the transition direction. To allow their precomputation, they can however be redefined to make them independent of the transition direction.
More precisely, the special node-type switching function $\beta^{(1)}_s$ (Equation~\ref{eq:special_node_beta2} and Figure~\ref{fig:special_switching1}) can be redefined as follows: 
\begin{equation}
  \bar\beta^{(1)}_s(v,x) =
  \begin{cases}
    \frac{1}{s} & \text{ if } \phi(v) \in \Sigma_{\phi_\mathcal{S}}\; \underline\vee\; \phi(x) \in \Sigma_{\phi_\mathcal{S}}\\
    \frac{1}{s^2} & \text{ if } \phi(v) \in \Sigma_{\phi_\mathcal{S}} \wedge \phi(x) \in \Sigma_{\phi_\mathcal{S}}\\
    1 & \text{otherwise},
  \end{cases}\hspace{0pt}
  \label{eq:impl_special_node_beta1}
\end{equation}

where $\underline\vee$ denotes the exclusive OR operator.
Similarly, we can redefine the function $\beta^{(2)}_s$ associated to the second strategy (Equation~\ref{eq:special_node_beta2} and Figure~\ref{fig:special_switching2}) as:

    \begin{equation}
      \bar\beta^{(2)}_s(v,x) =
      \begin{cases}
        \frac{1}{s} & \text{ if } \phi(v) \in \Sigma_{\phi_\mathcal{S}} \vee \phi(x) \in \Sigma_{\phi_\mathcal{S}}\\
        1 & \text{otherwise}.
      \end{cases} \hspace{0pt}
      \label{eq:impl_special_node_beta2}
    \end{equation}
%\smallskip

Figures~\ref{fig:special_switching_implementation1} and \ref{fig:special_switching_implementation2} depict the redefined functions for the two special node-type switching strategies.

These new definitions satisfy three key properties when transitioning from a node $v$ to any of its neighbors $x \in \mathcal{N}(v)$. 
For $k\in\{1,2\}$:
(1) If $v$ is a non-special node, then $\bar\beta^{(k)}_s(v,x)=\beta^{(k)}_s(v,x)$;
(2) If node $v$ is a special node, then $\bar\beta^{(k)}_s(v,x)=\tfrac1s\beta^{(k)}_s(v,x)$;
(3) $\bar\beta^{(k)}_s(v,x)=\bar\beta^{(k)}_s(x,v)$.
Properties (1) and (2) imply that the unnormalized transition probabilities and their implementations differ by a multiplicative constant. Thus the normalized transition probabilities induced by $\beta^{(k)}_s$ and $\bar\beta^{(k)}_s$ are the same. A proof of this fact is provided in section~\ref{sec:special_node_switching_comparison} of the Supplementary Information.
Finally, property (3) shows that the function $\bar\beta^{(1)}_s$ and  $\bar\beta^{(1)}_s$ depends only on the edge used by the \RW but not on the direction of the walk, thus the values of the function can be precomputed for every edge and considered as a multiplicative factor, similar to the weight of the edge. 

\begin{figure}[htbp]
  \centering
  \includegraphics[width=.75\linewidth]{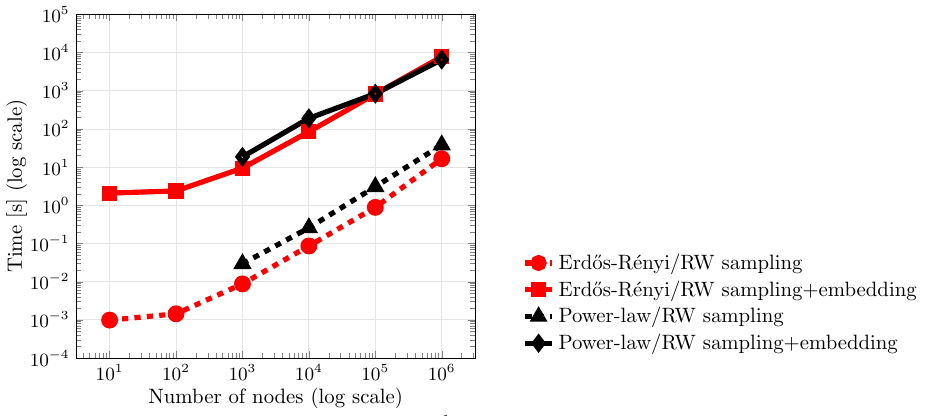}
  \caption{Computation time evolution of the \Het representation in Erdős-Rényi graphs with an
average degree of 10 and random power-law graphs with exponent $\alpha=2.2$.
For each family, graphs nodes and edges were randomly assigned to one of ten classes. 
Starting from each node, 10 random walks of length 100 were generated using the parameters $p=0.25, q=4$, and $s=0.5$. 
The set of \RW was used as the input of a Skipgram model to produce a vectorial embedding of size 100.
Computations have been performed using a processor AMD Rome 7452, 2.3 GHz, 32 cores with a RAM of 1024GB.
}
  \label{fig:erdos_renyi_time}
%\Description{Plot of the time employed by the different steps of Het-node2vec with respect to the size of the network for Erdős-Rényi and power-law random graphs. }
\end{figure}

The previous discussion ensures that using \Het with a special switching strategy does not increase complexity in either space or time compared to \ntv.
To confirm this claim, similarly to~\cite{Grover16}, 
we performed an empirical analysis of the computation time evolution using the generic strategy in \Het for both random-walk sampling process and sampling process plus the embedding construction (Figure~\ref{fig:erdos_renyi_time}).
The evaluation is conducted on two random graph models: the Erdős-Rényi model, where the edge probability is set to obtain a graph with average degree of ten; and a model generating random graphs with power-law degree distribution characterized by an exponent $\alpha=2.2$~\cite{VigerLatapy15}, which is a common graph topology in many practical applications.
As can be seen, in both models, the evolution of the computation time in the plot follows a linear trend, as in the case of \ntv~\cite{Grover16}.

%%%%%%%%%%%%%%%%%%%%%%%%%%%%%%%%%%%%%%
\section{Experimental results}
\label{sec:exp-res}
To assess the effectiveness of the representations generated by \Het we performed two sets of experiments.
We firstly designed experiments to study the influence of the switching parameters on \Het performance, using both synthetic (Section~\ref{sec:comparison_synthetic_data}) and real benchmark data sets (Section~\ref{sub:analysis-switch-real}).
Then, we compared \Het~with popular state-of-the-art HGRL methods using the heterogeneous graph benchmark framework proposed in~\cite{yang2020heterogeneous} on standard supervised learning tasks: node label classification  and edge prediction (Section~\ref{sec:comparison_alternative_models}).

%%%%%%%%%%%%%%%%%%%%%%%%%%%%%%%%%%%%%%%%%%%%%%%%%%%%%%%%%%%%%
\subsection{Analysis of the  switching parameters of \Het~using  synthetic data sets}
\label{sec:comparison_synthetic_data}
We observe that \ntv\ is a particular case of \Het\ when $s=1$. For this reason we compared the behaviour of \Het\ with \ntv\, for different values of the $s$ parameter.

\begin{figure}[htbp]
   \centering
   \begin{subfigure}[b]{\linewidth}
     \centering
     \includegraphics[width=0.3\linewidth]{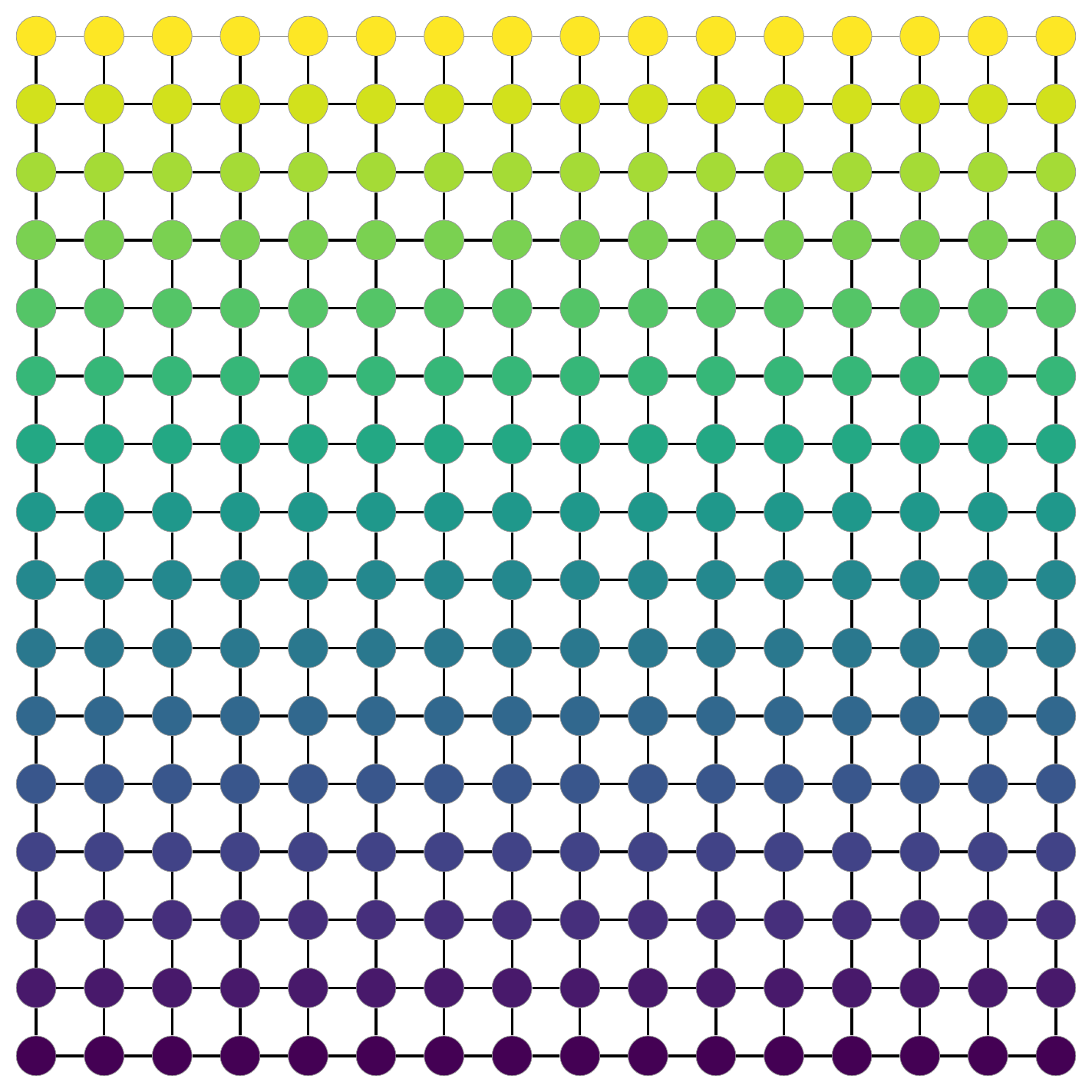}

     \caption{}
     \label{fig:grid_topology}
   \end{subfigure}
   \hfill
   \begin{subfigure}[b]{\textwidth}
     \includegraphics[width=1.1\textwidth]{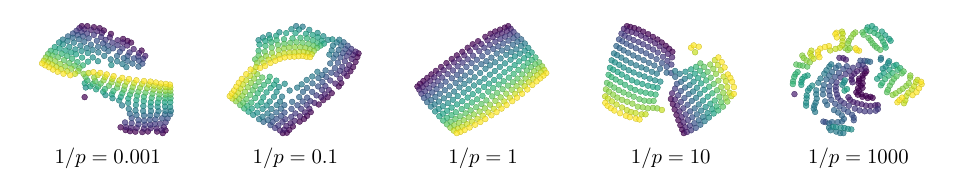}
     \includegraphics[width=1.1\textwidth]{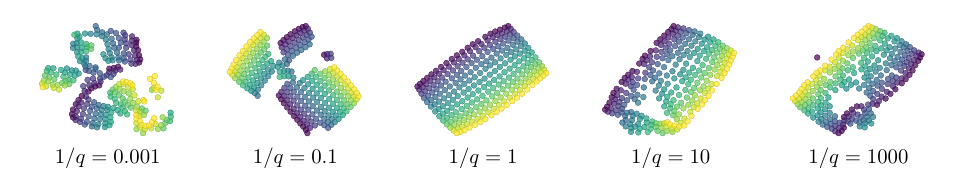}
     \caption{}
     \label{fig:grid_size_10_p_TSNE}
   \end{subfigure}
   \hfill
   \begin{subfigure}[b]{\textwidth}
     \includegraphics[width=1.1\textwidth]{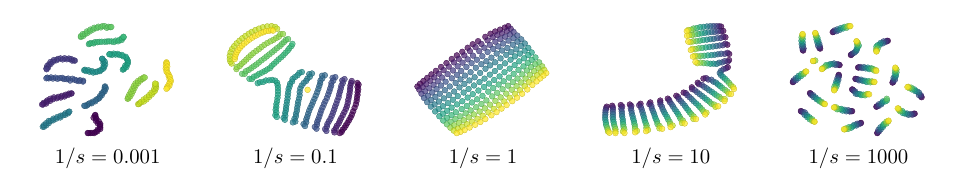}
     \caption{}
   \label{fig:grid_size_10_s_TSNE}
   \end{subfigure}
   \caption{
   Comparison of \Het~and \ntv embeddings on a synthetic heterogeneous graph having different types of nodes.
   (a) Heterogeneous synthetic grid-shaped graph. Node types are represented through different colors, and nodes in the same row belong to the same type.  For each setting of the $s$, $q$ and $p$ parameters in \Het, embeddings are computed by generating 10 RWs of length 128 from each node. These paths are input to a Skipgram architecture, with a window size equal to 5, to generate an embedding of the nodes into a 10D space. The final 2D representation is obtained by t-SNE~\cite{vanderMaaten08}.
   (b) \ntv~embeddings of the grid from different inward switching values $p$ and $q=s=1$ (first row); in-out parameters  $q$ and $p=s=1$ (second row).
   (c) \Het~embeddings of the grid for different node type switching values $s$ and $p=q=1$.
 }
   \label{fig:embedding_s}
 \end{figure}

In Figure~\ref{fig:embedding_s}, we present an example of the application of \Het on a synthetic grid topology where edges are unweighted and homogeneous, while nodes have heterogeneous types. 
Each grid row corresponds to a different node type, depicted with different colors in the figure.
The different plots show the 2-D  t-SNE projections~\cite{vanderMaaten08} of the embeddings obtained changing the values of the hyperparameters of \ntv and \Het.
  In Figure~\ref{fig:grid_size_10_p_TSNE} are depicted the embeddings obtained for different values of $p$ ($q=1$) and $q$ ($p=1$) in \ntv; Figure~\ref{fig:grid_size_10_s_TSNE} shows the influence of the switching parameter $s$ ($p = 1$, $q = 1$) in the generic node-type switching of \Het.

% We notice the clear impact of the node-switching hyperpameters on the final embedding, specially for the node-type switching parameter.
As expected, \ntv ($1/s=1$) captures the topological characteristics of the grid, but it is unable, even for different values of $p$, to cluster together nodes of the same type (Figure~\ref{fig:grid_size_10_p_TSNE}).
% When the values of $s$ are such that the note-type switch is discouraged ($1/s<1$) the computed embeddings cluster nodes with the same types.  
 On the other hand, by properly tuning the switching parameter $s$, we can obtain a graph embedding that is aware of the node types. Small values of the node-type switching probability $\beta_s$ (obtained when $1/s<1$, e.g. $1/s=10^{-3}$ - Figure~\ref{fig:embedding_s}c) tend to cluster nodes according to their type - i.e., nodes belonging to the same row are indeed clustered,
 while large node-type switching probabilities $\beta_s$ ($1/s=10^3$) gather together the grid columns.

Setting intermediate values of $s$ has the effect of discouraging or increasing the switching probability, still allowing some inter-type and intra-type walks. This effect can be observed in Figure~\ref{fig:grid_size_10_s_TSNE} for values of $1/s=0.1$ and $1/s=10$. When $1/s=0.1$, switching is discouraged, leading to nodes of the same type being embedded closely together, forming type-specific clusters. However, a few type-switches are still permitted, resulting in clusters that group rows adjacent in the grid. On the other hand, when $1/s=10$, the probability of moving between node-types (rows) increases, leaving room for intra-type visits; we observe clusters containing nodes with completely different types (each cluster contains nodes from one column of the grid), while clusters that are close to each other contain nodes from neighboring columns.
When $\frac{1}{s}=1000$ the probability of switching is in practice close to $1$ and clusters are composed by nodes of different types according to the columns of the grid, and also the topological closeness of the columns is lost, since in practice no inter-column moving is allowed.

Note that the set of representations obtained in \ntv using $p=1$ are similar to those using ($q=1$) but in inverse order. 
  This behaviour is explained by the fact that the graph in the example does not contain any triangle (a complete subgraph of three vertices); therefore, the transition probabilities during the generation of a RW can only take the values $1/p$ or $1/q$.
Therefore, the ratio $q/p$ fully characterizes the parameter space by quantifying the relation between the local/global type of the generated paths.
A broader comparison of parameter sensitivity can be found in the Supplementary Information (section~\ref{sec:appendix_grid_embedding}), where different embeddings are constructed by varying parameters $p$, $q$, and $s$.

 We further analyzed the impact of the $s$ switching parameter using another synthetic data set (Figure~\ref{fig:clustered-grid} of section~\ref{supp:switching-exp} in the Supplementary Information). In that figure, it is shown the embeddings obtained on a synthetic graph composed of nine node types, organized into $3 \times 3$ square-grid, where each square is composed of nodes sharing the same type.
 The embeddings are computed using \Het with $p = 1$, $q = 1$, and by setting different values for the generic switching parameter $1/s$ (Supplementary Information -- Figure~\ref{fig:clusterd_grid_size_10_p_TSNE}).
 It can be seen that by discouraging the node type switching ($1/s\ll 1$), we obtain embeddings that tend to cluster the nodes belonging to the same type.
A similar example in a real-world graph is depicted in Figure~\ref{fig:cora_size_10_p_TSNE} of section~\ref{supp:switching-exp} where different embedding are generated for the well-known Cora network \cite{grape23} representing the relation between scientific publications.
  Both examples show that the t-SNE projections of \Het-embeddings can separate the different node types more effectively.

\subsection{Analysis of the  switching parameters of \Het on real benchmark data sets}
\label{sub:analysis-switch-real}
In this section, we analyze the impact of the switching parameters in the graph representation obtained with \Het~using the datasets and the experimental set-up proposed in~\cite{yang2020heterogeneous}. 
The datasets used in the experiments are described in Section~\ref{sec:datasets}, and the experimental set-up in Section~\ref{sec:exp_settings}. The same framework has also been used to compare \Het~with state-of-the-art HGRL methods (Section~\ref{sec:comparison_alternative_models}).

%%%%%%%%%%%%%%%%%%%%%%%%%%%%%%%%%%
\subsubsection{Datasets/Graphs}
\label{sec:datasets}

In our experiments, we utilize four heterogeneous graphs proposed in in~\cite{yang2020heterogeneous} as benchmark for the comparison of HGRL methods, i.e. Freebase, DBLP, YELP, and PubMed.

The differences among the provided graphs extend beyond node and edge cardinalities and include node-type/edge-type distribution, node-degree distribution, and graph connectivity.
Consequently, we believe that evaluating \Het using these datasets allows us to assess its robustness and generalizability across different scenarios.

Table~\ref{tab:graphs_macro} depicts the macroscopic features of the graphs.
As can be noticed, the graph dimensions and the number of different node/edge types span a large range of values.

In the Supplementary Information, we provide more details about the characteristics of the four graphs: section~\ref{app:datasets}  includes the node type distribution and basic degree statistics, depicted in Table~\ref{tab:node_type_distribution}; the edge type distribution, summarized in Figure~\ref{fig:edge_type_distribution}; and the complementary cumulative distribution function (CCDF) of node degrees in Figure~\ref{fig:ccdf}.

%%%%%%%%%%%%%%%%%%%%%%%%%%%%%%%%%%%%%%%%%%
\begin{table}[htb]
  \caption{Macroscopic description of the four heterogeneous network datasets.}
  \label{tab:graphs_macro}
  \centering
%  \resizebox{.8\linewidth}{!}{
    \begin{tabular}{lrrr rrr }
    \toprule
      Dataset         & \# nodes   & \# edges    & mean degree & \multicolumn{3}{c}{\# types}    
      \\
              \cmidrule(l){5-7}    
                      &            &             &              & nodes & edges & labels     \\%
    \hline   Freebase & 12,164,758 & 62,982,566  & 10.35       & 8 & 36    & 8               \\
    DBLP     & 1,989,077  & 258,850,593
    & 277.46      & 4 & 6     & 13                         \\
    Yelp     & 82,465     & 30,542,675  & 740.74      & 4 & 4     & 16                        \\
    PubMed   & 63,109     & 244,986     & 7.76       & 4 & 10    & 8                          \\
    \bottomrule
    \end{tabular}
%  }
\end{table}

%%%%%%%%%%%%%%%%%%%%%%%%%%%%%%%%%%%%%%%%%%%%%%%%%%%%%%%%%%
\subsubsection{Experimental settings}
\label{sec:exp_settings}

Following the experimental setting proposed in~\cite{yang2020heterogeneous}, we applied \Het\ in an unsupervised and unattributed setting to assess the quality of the computed representations. We embedded the four benchmark graphs (Section \ref{sec:datasets}) and used the computed embeddings to perform, on each graph, the two predictive tasks proposed by the benchmark framework: node-label prediction and edge-prediction tasks on specific node and edge types.
Details about the characteristics of the prediction tasks considered in the experiments are available in the Supplementary Information Section~\ref{suppl:pred-tasks}.

\begin{figure}[htb]
  \centering
  \includegraphics[width=\linewidth]{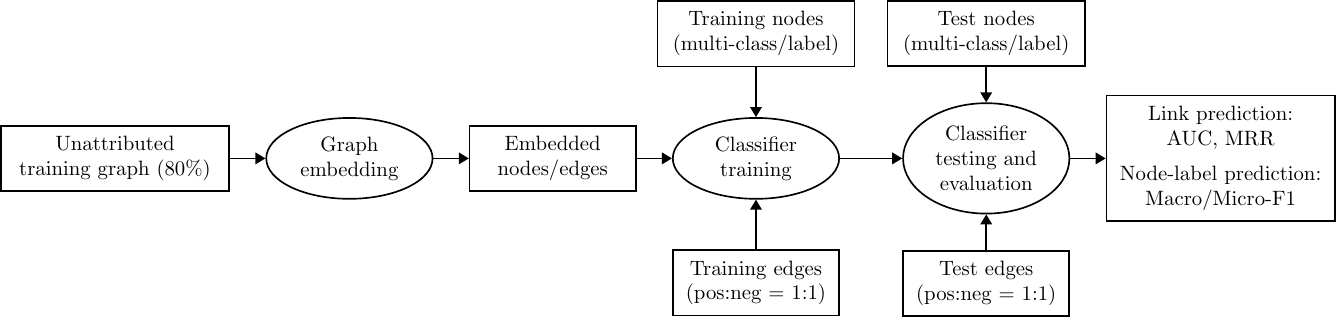}
  \caption{HNE experimental evaluation workflow.}
  %\EC{Bottom: Ho messo una alternativa sotto} 
  \label{fig:HNE_schema}
  %\Description{Diagram of the evaluation workflow used in the experimental section.}
\end{figure}

\paragraph{Evaluation workflow.}
The workflow of the evaluation pipeline is depicted in Figure~\ref{fig:HNE_schema}. The process begins with the removal of 20\% of existing edges in a random fashion to obtain an independent test set for unbiased evaluation. The remaining dataset is used to compute a vectorial representation of the graph nodes according to the different embedding methods. 

According to the experimental setting used in \cite{yang2020heterogeneous}, the embedding is computed in an unsupervised setting, attempting to minimize a loss function that preserves the graph-topological structure. 
When used under this specific setting, GNNs are trained to maximize the likelihood of observing edges in the heterogeneous
network. To this aim, they used negative sampling and minimized the cross-entropy loss (see the code provided by authors of \cite{yang2020heterogeneous}).

The embedding obtained from the previous step serves as the resulting representation for the subsequent predictive tasks.
According to~\cite{yang2020heterogeneous}, the embedding size is set to 50, while other crucial parameters are optimized on all datasets through standard five-fold cross-validation on the training graph\footnote{To apply the benchmarked algorithms, Yang et al.  in~\cite{yang2020heterogeneous} adapted the code provided by the authors of the algorithms themselves. The parameters that are optimized therefore depend on the code itself, as chosen by the authors of the code.}.

Node-label classification is performed by training and evaluating a linear support vector machine~\cite{fan2008liblinear} on a subset of labeled nodes using five stratified holdouts (80:20 train:test ratio).   

For the edge-prediction task, 20\% of left-out edges are integrated with an equal amount of randomly sampled negative edges (i.e., non-existing edges) of the same type; this creates a balanced set of positive and negative edges. Next, the Hadamard operator is applied to the embeddings of the source and destination nodes to obtain the edge embeddings. 

The edge-prediction task is then carried out using five-holdout cross-validation (80:20 train:test ratio) and linear support vector machines as classifier models.
The evaluation is conducted by averaging the AUC (area under the ROC curve) and the MRR (mean reciprocal rank) performance measures obtained on each holdout. While AUC is a standard measure for classification when link prediction is regarded as a binary (existence/non-existence) classification problem, MRR is a standard measure for ranking when link prediction is used for link retrieval. Formally, if $V_{test}$ is the set of source nodes of the test edges, the MRR is computed as 
$$
\text{MRR} = \frac{1}{|V_{test}|} \sum_{v \in V_{test}} \frac{1}{|E_{+}|} \sum_{e \in E_{+}} \frac{1}{\text{rank}_e},
$$
where $E_{+}$ is the set of positive edges starting from the source node $v \in V_{test}$, and $\text{rank}_e$ is the rank of the value computed by the svm for the positive test edge $e$ with respect to all edges starting from $v$ (both negative and positive). In practice, the MRR assesses the model's ability to assign high scores to true edges and low scores to negative edges.

For the node-label classification task, the performance measures averaged across the 5 holdouts are the macro-F1 (across all labels) and micro-F1 (across all nodes). 
To guarantee a fair and repeatable comparison, we perform the predictive node-label and edge-prediction tasks by using the training/test splits provided by authors of~\cite{yang2020heterogeneous} as well as their implementation of the evaluation pipeline. %provided by the authors of~\cite{yang2020heterogeneous},  The data provided by the benchmark contain the training/test splits. %; in more detail, a seed is used to guarantee that the holdout splits will be the same across all the evaluated models. In this way, authors of ~\cite{yang2020heterogeneous} ensure the comparability and repeatability of the results.

\paragraph{\Het hyperparameter settings.}
\label{sec:het_hyper}

A fair comparison of \Het against its homogeneous counterpart, \ntv, can be obtained by fixing the values of the $p$ and $q$ hyperparameters, ensuring a consistent node-neighborhood sampling strategy (DFS-like or BFS-like visit) across all the graphs. With this aim, we conducted a preliminary evaluation using internal five-fold cross-validation on the two smallest graphs, PubMed and Yelp.

Specifically, we experimented with values of $p$ and $q$ in the set $\{0.25, 0.5, 0.1, 4, 10, 100\}$. We found that the most robust and accurate results were achieved by setting $1/p = 0.25$ and $1/q = 4$. These values were kept fixed across all experiments on all benchmark graphs, biasing all RWs towards a DFS-like visit that explores new regions of the graph.

For each graph, the number of RWs starting from each node was chosen between 10 and 50, based on internal three-fold cross-validation. These values were 
%MM chosen 
identified by balancing computational time against the representativeness of the sampled node contexts. A high number of RWs yields embeddings that better represent the topological relationships in the graph but increases computational time, while a low number of RWs ensures faster computation but at the expense of embedding accuracy. 

We used a skipgram mode as the shallow neural network for embedding, with the network hyperparameters values set to their default values according to the implementation provided by GRAPE (see Table \ref{tab:grape_skipgram} in 
Section~\ref{sec:hyperparameters} of the Supplementary Information). 
The previous procedure led to the vectorial representation of the network nodes.

We empirically evaluated the results obtained from \Het using either the generic node-type-switching strategy or the special node-type-switching strategy (Section~\ref{subsub:generic-multiple-special-switch}), where the special node-type is set as the node targeted by the node-label prediction task (see Table~\ref{tab:node_type_distribution} in the Supplementary Information). 
Notably, for all benchmark graphs except Yelp, the node-type targeted by the node-label prediction is also the type characterizing the endpoints of the edges targeted by the link-prediction task. In this case, applying the special node-type switch can also benefit edge prediction by increasing the chances of stepping through the edge of interest. 
For this reason, we used special node type switching also for the edge prediction task.

%%%%%%%%%%%%%%%%%%%%%%%%
\subsubsection{\Het~sensitivity on switching parameter}
\label{sec:sensitivity_switching_parameter}

In this section, we empirically evaluate the relationship between the switching parameter in \Het and the performance obtained in the predictive tasks. To this aim, and to avoid confounding effects, we only consider the node type heterogeneity. We conduct different experiments by fixing the return parameter $1/p=0.25$, the outward parameter $1/q=4$, and then changing the value of the (generic or special) node-type switching parameter $\beta_s$ in the range $[10^{-1}, 10^{2}]$. This analysis attempts to assess whether the use of the semantic information induced by node types could impact the quality of the obtained graph representation.

\begin{figure}[htbp]
  \centering
    \begin{subfigure}{0.45\textwidth} \centering
      \resizebox{.75\linewidth}{!}{
      \includegraphics[width=\linewidth]{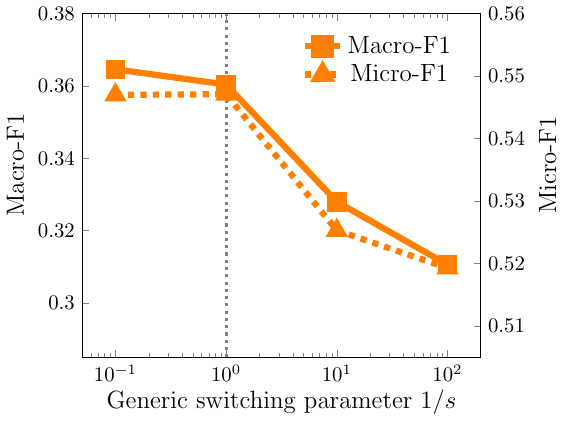}}
    \caption{Freebase generic}
  \end{subfigure}
   \begin{subfigure}{0.45\textwidth} \centering
     \resizebox{.75\linewidth}{!}{
       \includegraphics[width=\linewidth]{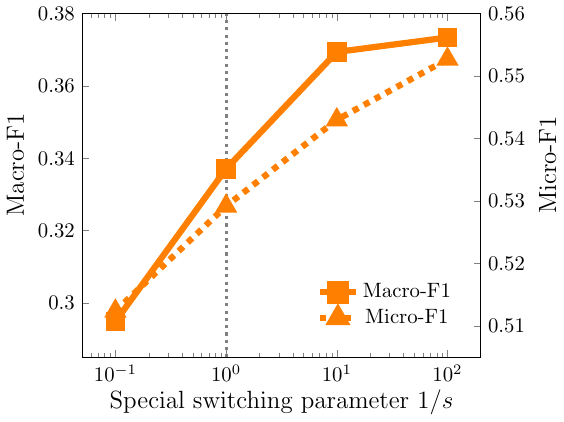}}
    \caption{Freebase special}
  \end{subfigure}

  \begin{subfigure}{0.45\textwidth} \centering
    \resizebox{.75\linewidth}{!}{
      \includegraphics[width=\linewidth]{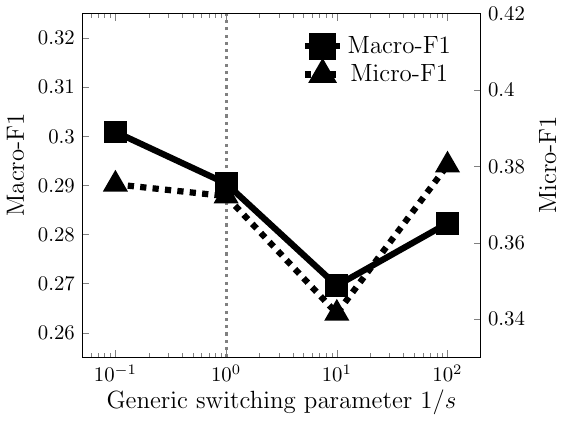}}
    \caption{DBLP generic}
  \end{subfigure}
    \begin{subfigure}{0.45\textwidth} \centering
      \resizebox{.75\linewidth}{!}{
      \includegraphics[width=0.8\linewidth]{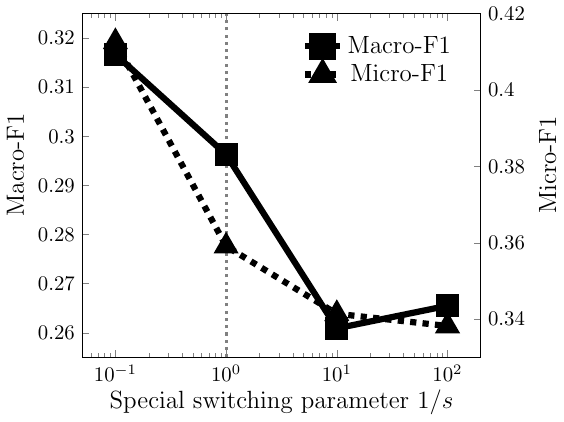}}
    \caption{DBLP special}
  \end{subfigure}

  \begin{subfigure}{0.45\textwidth} \centering
    \resizebox{.755\linewidth}{!}{
    \includegraphics[width=\linewidth]{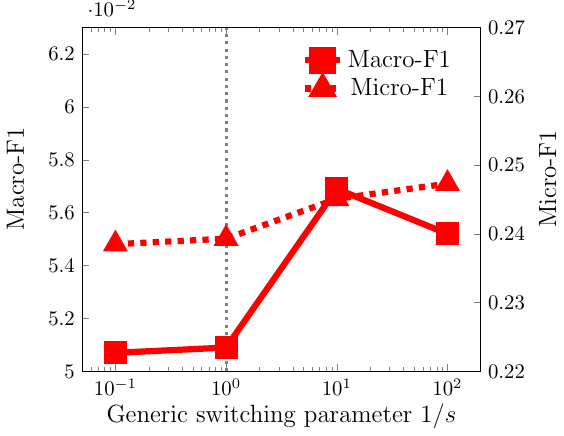}}
    \caption{Yelp generic}
  \end{subfigure}
  \begin{subfigure}{0.45\textwidth} \centering
    \resizebox{.75\linewidth}{!}{
      \includegraphics[width=\linewidth]{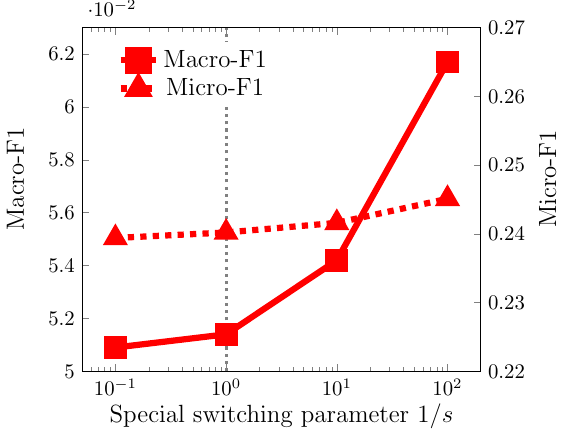}}
    \caption{Yelp special}
  \end{subfigure}

  \begin{subfigure}{0.45\textwidth} \centering
    \resizebox{.75\linewidth}{!}{
      \includegraphics[width=\linewidth]{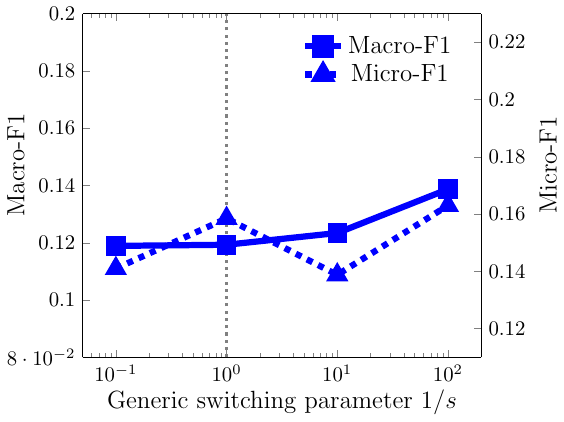}}
    \caption{Pubmed generic}
  \end{subfigure}%
  \begin{subfigure}{0.45\textwidth} \centering
    \resizebox{.75\linewidth}{!}{
      \includegraphics[width=\linewidth]{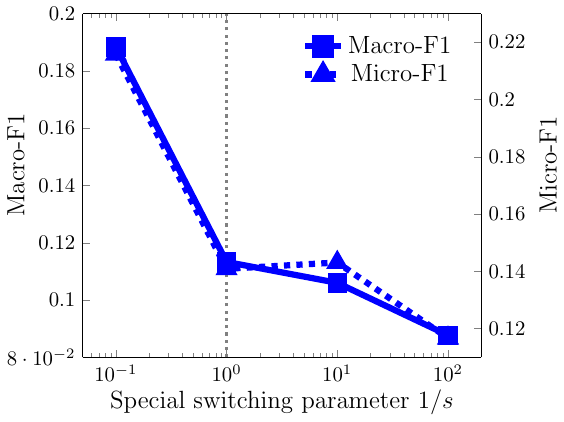}}
    \caption{Pubmed special}
  \end{subfigure}%

  \caption{Node-label prediction: macro-F1 (left axis - continuous line) and micro-F1 (right-axis - dashed line) values obtained for varying values of the $\beta_s$ parameter.
    Plots a,c,e,g in the first column depict the variation of the performance metrics when the generic switching strategy is used; the second column (subplots b,d,f,h) shows the results obtained when using the special switching strategy. In each plot, different scales are used for the left (macro-F1) and right (micro-F1) axis. The plots referring to the same graph use the same macro-F1 and micro-F1 scales to allow a comparison between the generic and the special node-type switching strategy.}
  \label{fig:node_label_prediction_s}
%\Description{Plots of the evolution of the quality metrics for the node-label prediction task on all the benchmark datasets with respect to increasing values of the switching parameter in the generic and special strategies.}
\end{figure}

\paragraph{Node label prediction.}
\label{sec:sensitivity_node_label}
Figure~\ref{fig:node_label_prediction_s} demonstrates how varying the value of the switching parameter $s$ affects the Macro / Micro-F1 values achieved.
  This effect is observed when both the generic (Figure~\ref{fig:node_label_prediction_s} a,c,e,g) and the special (Figure~\ref{fig:node_label_prediction_s} b,d,f,h) node-type switching strategies are employed.

With the generic switching strategy (Figure~\ref{fig:node_label_prediction_s}, left column), both Freebase and DBLP exhibit a decline in performance as the switching probability $\beta_s = 1/s$ increases. In contrast, Yelp and PubMed display an opposite trend. Special switching behaves differently, showing a positive correlation between $1/s$ and the $F1$ scores for Freebase and Yelp, and a negative correlation for DBLP and PubMed (Figure~\ref{fig:node_label_prediction_s} b, d, f and h).

This behavior can be explained by the fact that in both Freebase and DBLP, labeled nodes (of type \texttt{book} in Freebase and type \texttt{author} in DBLP) have lower degrees compared to other node types (see Figures~\ref{fig:ccdf_freebase} and \ref{fig:ccdf_DBLP}).
In these cases, generic node-type switching could cause the RW to move away from the nodes of interest, thus failing to represent their topological neighborhoods (as shown by the decreasing trends in Figures~\ref{fig:node_label_prediction_s}a and \ref{fig:node_label_prediction_s}c).
This is particularly true in Freebase, where only approximately 23\% of nodes are of type \texttt{book} (Table~\ref{tab:node_type_distribution} in the Supplementary Information); using the special node-type switching strategy to bias the walk in favor of \texttt{book} nodes (i.e., using high values of $\beta_s$, see Figure~\ref{fig:node_label_prediction_s}e) achieves better results, as indicated by the increasing trend.
Conversely, in DBLP, \texttt{author} nodes, which account for approximately 89\% of all nodes, are the most represented node type. Here, using the special node-type switching strategy with high values of $\beta_s$ would confine the walk to \texttt{author} nodes, neglecting other node types and resulting in a loss of information in the computed embedding. This is evident in Figure~\ref{fig:node_label_prediction_s}d, where performance decreases as $\beta_s$ increases.

In the Yelp graph (Figures~\ref{fig:node_label_prediction_s}e and \ref{fig:node_label_prediction_s}f), the node-label prediction focuses on \texttt{business} nodes, which are much less represented (approximately 9\% of all nodes) compared to \texttt{phrase} nodes, which dominate the graph, constituting approximately 91\% of all nodes (Table~\ref{tab:node_type_distribution} in the Supplementary Information). In this scenario, using a generic type-switching strategy that enforces heterogeneity (high values of $\beta_s$) has roughly the same positive effect as a special node-type switching strategy where the special node type is \texttt{business}; in both cases, high values of $\beta_s$ promote visits to business nodes. Both settings yield better performance for high values of $1/s$ (Figures~\ref{fig:node_label_prediction_s}e and \ref{fig:node_label_prediction_s}f) because they compel the walk to move away from phrase nodes and capture as much heterogeneous information as possible.

PubMed is characterized by the least unequal node-type distribution and edge-degree distribution (Table~\ref{tab:node_type_distribution}). In other words, with the exception of \texttt{species} nodes, the types of nodes in the PubMed network are more evenly distributed than in the other benchmark graphs, with no overwhelming majority of any node type. Additionally, the degree distributions are nearly identical for all node types (see Figure~\ref{fig:ccdf_pubmed} in the Supplementary Information). In this case, favoring the heterogeneity of the walk has only a minor positive effect on performance (slightly increasing trend in Figure~\ref{fig:node_label_prediction_s}g), and using a special node-type switch that forces the walk toward \texttt{disease} nodes (the nodes used in the node-label prediction task) leads to walks that focus on \texttt{disease} nodes (Figure~\ref{fig:node_label_prediction_s}h).

The results presented in this section highlight that the relative distribution of node types and their degrees must be accounted for when setting the switching parameter, as expected.
Moreover, a careful setting of this parameter improves the prediction performance of classic homogeneous RW-based embedding techniques.

\begin{figure}[htbp]
  \centering 
  \begin{subfigure}{0.45\textwidth} \centering
    \resizebox{.75\linewidth}{!}{
            \includegraphics[width=\linewidth]{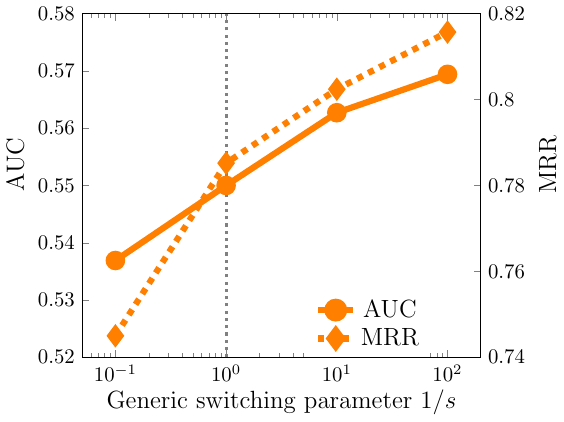}}
    \caption{Freebase generic}
  \end{subfigure}
    \begin{subfigure}{0.45\textwidth} \centering
      \resizebox{.75\linewidth}{!}{
        \includegraphics[width=\linewidth]{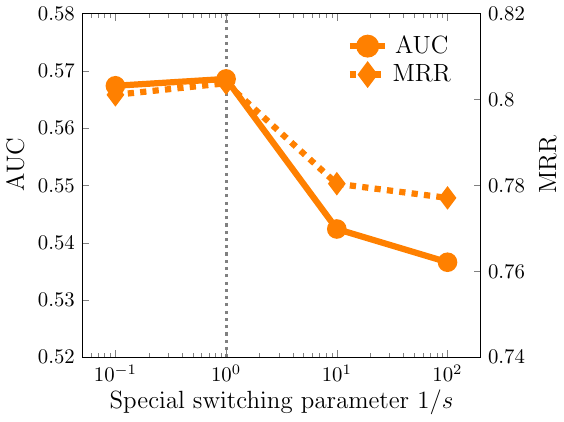}}
    \caption{Freebase special}
  \end{subfigure}

  \begin{subfigure}{0.45\textwidth} \centering
    \resizebox{.75\linewidth}{!}{
      \includegraphics[width=\linewidth]{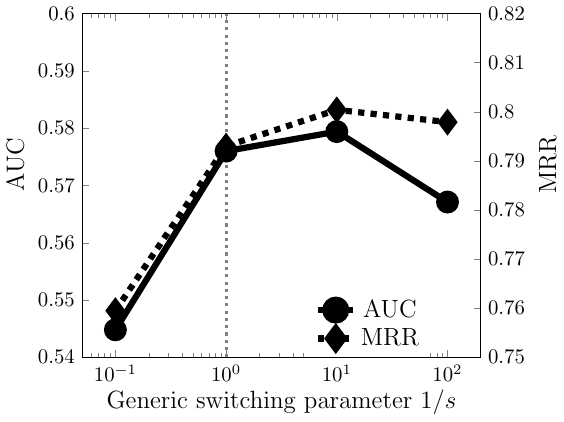}}
    \caption{DBLP generic}
  \end{subfigure}
  \begin{subfigure}{0.45\textwidth} \centering
    \resizebox{.75\linewidth}{!}{
      \includegraphics[width=\linewidth]{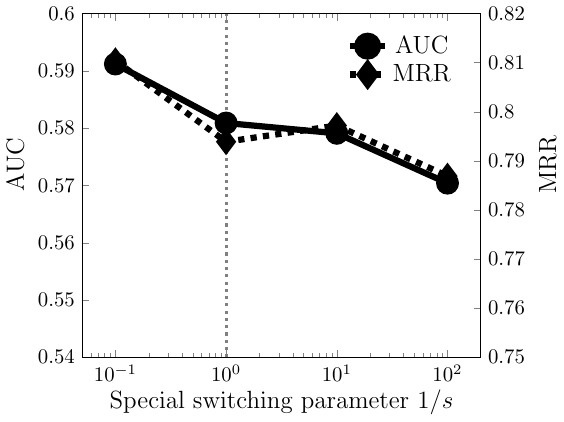}}
    \caption{DBLP special}
  \end{subfigure}

  \begin{subfigure}{0.45\textwidth} \centering
    \resizebox{.75\linewidth}{!}{
      \includegraphics[width=\linewidth]{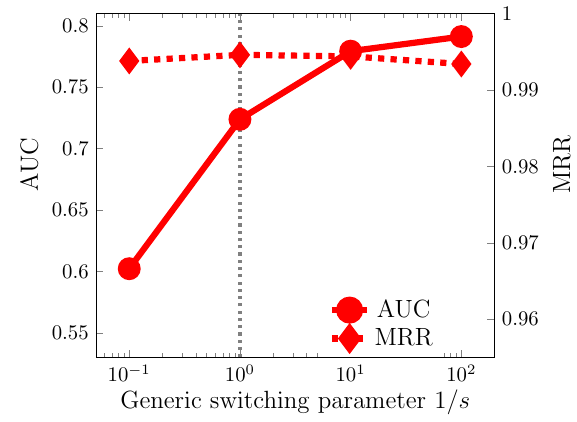}}
    \caption{Yelp generic}
  \end{subfigure}
  \begin{subfigure}{0.45\textwidth} \centering
    \resizebox{.75\linewidth}{!}{
      \includegraphics[width=\linewidth]{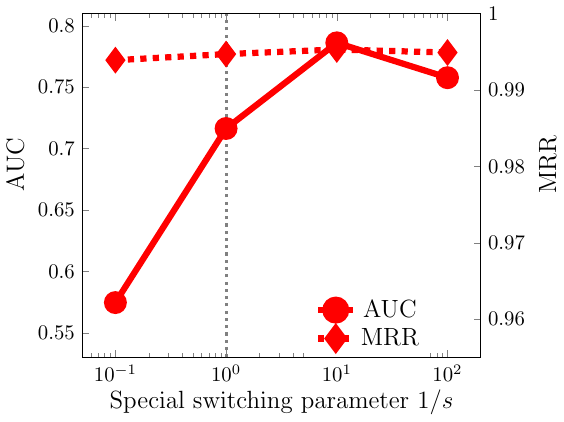}}
    \caption{Yelp special}
  \end{subfigure}

  \begin{subfigure}{0.45\textwidth} \centering
    \resizebox{.75\linewidth}{!}{
      \includegraphics[width=\linewidth]{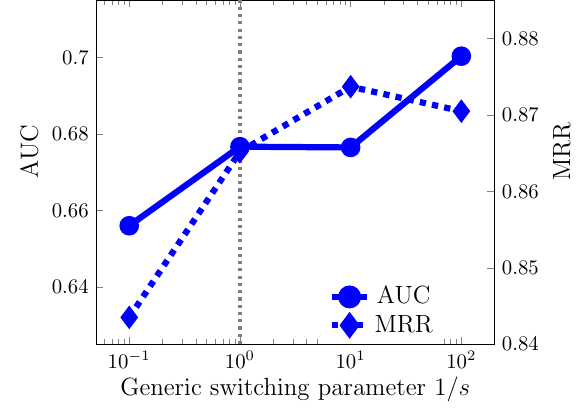}}
    \caption{Pubmed generic}
  \end{subfigure}
  \begin{subfigure}{0.45\textwidth} \centering
    \resizebox{.75\linewidth}{!}{
      \includegraphics[width=\linewidth]{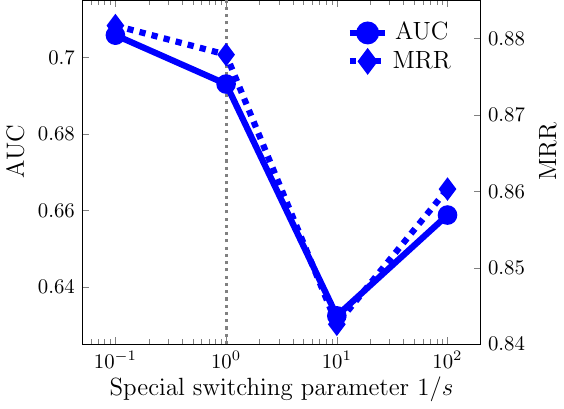}}
    \caption{Pubmed special}
  \end{subfigure}%

  \caption{Edge prediction: AUC (left axis - continuous line) and MRR (right-axis - dashed line) values obtained for varying values of the $\gamma_c$ parameter.
    Plots a,c,e,g in the first column depict the variation of the performance metrics when the generic switching strategy is used; the second column (subplots b,d,f,h) shows the results obtained when using the special switching strategy. In each plot, different scales are used for the left (AUC) and right (MRR) axis. The plots referring to the same graph use the same AUC and MRR scales to allow a comparison between the generic and the special node-type switching strategy.}
  \label{fig:link_prediction_s}
%%\Description{Plots of the evolution of the quality metrics for the edge prediction task on all the benchmark datasets with respect to the switching parameters in the generic and special strategies.}

\end{figure}

\paragraph{Edge prediction}
\label{sec:sensitivity_edge_predicition}
The results obtained in the edge-prediction task, depicted in Figure~\ref{fig:link_prediction_s}, provide further evidence that the proposed type-aware RWs improve node2vec results.

When observing results achieved by the generic-type switching strategy (left column in Figure~\ref{fig:link_prediction_s}), better results are always achieved when a higher value of the switching parameter is used. In other words, promoting the heterogeneity of the node-types, i.e., biasing the walk to step through edges with endpoints of different types, correlates with an improved graph representation. This is due to a better exploration of the heterogeneous context of nodes at the endpoints of the targeted edges (highlighted in gray in Figure~\ref{fig:edge_type_distribution} in the Supplementary Information).

When using the special node-type switching strategy (right column of Figure~\ref{fig:link_prediction_s}), we note an opposite decreasing trend for all graphs with the exception of Yelp. For Freebase, DBLP, and PubMed, the types of edges targeted by the edge-prediction task, as well as the node types at their endpoints, are well represented in the graph. Using high $\beta_s$ values to promote switching to special node types (i.e., the node types at the endpoints of the targeted edges) results in almost homogeneous RWs that repeatedly visit only special nodes and use only the targeted edges, neglecting others. This results in embeddings that are not sufficiently informative.

In Yelp, we instead observe an increasing trend mainly because the edges targeted by the link prediction (\texttt{business\--described-with-phrase}, see Figure~\ref{fig:edge_type_distribution} in the Supplementary Information) connect the special node type \texttt{business}, which is relatively rare (approximately 9\% of all nodes in the graph), to the non-special node type \texttt{phrase} which dominates the graph (approximately 91\% of all nodes). As a result, the edges targeted by the link prediction are much less represented (approximately 9\% of all edges) than the most common \texttt{phrase-context-phrase} edges (approximately 91\% of all edges). Focusing the walk on \texttt{business} nodes enhances the likelihood of traversing the edge of interest, thereby producing embeddings that are informative for link prediction.

%\smallskip
Summarizing, we observe that for both the node label and edge prediction tasks the switching process that characterizes \Het~with respect to the classical \ntv algorithm ($1/s=10^0$, vertical dotted lines in Figure~\ref{fig:node_label_prediction_s} and \ref{fig:link_prediction_s}) improves the prediction results.

%%%%%%%%%%%%%%%%%%%%%%%%%%%%%%%%%%%%%%%%%%%%%

\subsection{Comparison of \Het~ with state-of-the-art HGRL methods}
\label{sec:comparison_alternative_models}

To show the effectiveness of \Het with respect to state-of-the-art techniques, we applied the same experimental setting published in \cite{yang2020heterogeneous} and described in Section~\ref{sec:exp_settings} to allow a fair comparison with:
\begin{itemize}
    \item five RW-based embedding methods (metapath2vec \cite{Dong17}, PTE \cite{Tang15}, Aspem \cite{shi2018aspem}, HIN2Vec \cite{fu2017hin2vec}, HEER \cite{shi2018easing});
    \item four GCN-based embedding methods (R-GCN \cite{Schlichtkrull18} and HAN \cite{Wang19bHAN}, HGT \cite{Hu20}, and MAGNN \cite{fu2020magnn});
    \item four Relational learning neural methods (TransE \cite{bordes2011learning}, DistMult \cite{Yang2014EmbeddingEA}, ConvE \cite{dettmers2018convolutional}, and ComplEx \cite{trouillon2016complex}).
\end{itemize}

%\paragraph{Node label prediction}
%\label{sec:node_label_prediction_results}

The results for the node-label prediction task shown in Table~\ref{tab:node_prediction_results}
 evidence the superiority of \Het in the node-label prediction for all graphs but DBLP. This indicates that the proposed heterogeneous RW approach effectively captures the underlying structure and semantic information of the graphs.

Special node type switching achieves the best results, but also generic type switching achieves, on average, better results than the other competing HGRL methods.
By tuning the $1/s$ parameter, both strategies lead to results that are better or comparable with state-of-the-art competing methods.

\begin{table}[htb]
\caption{Performance metric for the node label prediction task on the benchmark graphs. \textit{\Het-special} refers to the special node-type switching strategy, where the special node-type is the one targeted by the node-label prediction task. In the table, for both the \Het settings we report only the values obtained by the two extreme values of the node-type switching parameter $1/s = {0.1, 100}$. To avoid confounding effects, the edge-type switching parameter is set to $1/c = 1$.
  }
  \label{tab:node_prediction_results}
  \centering
  \includegraphics[width=\textwidth]{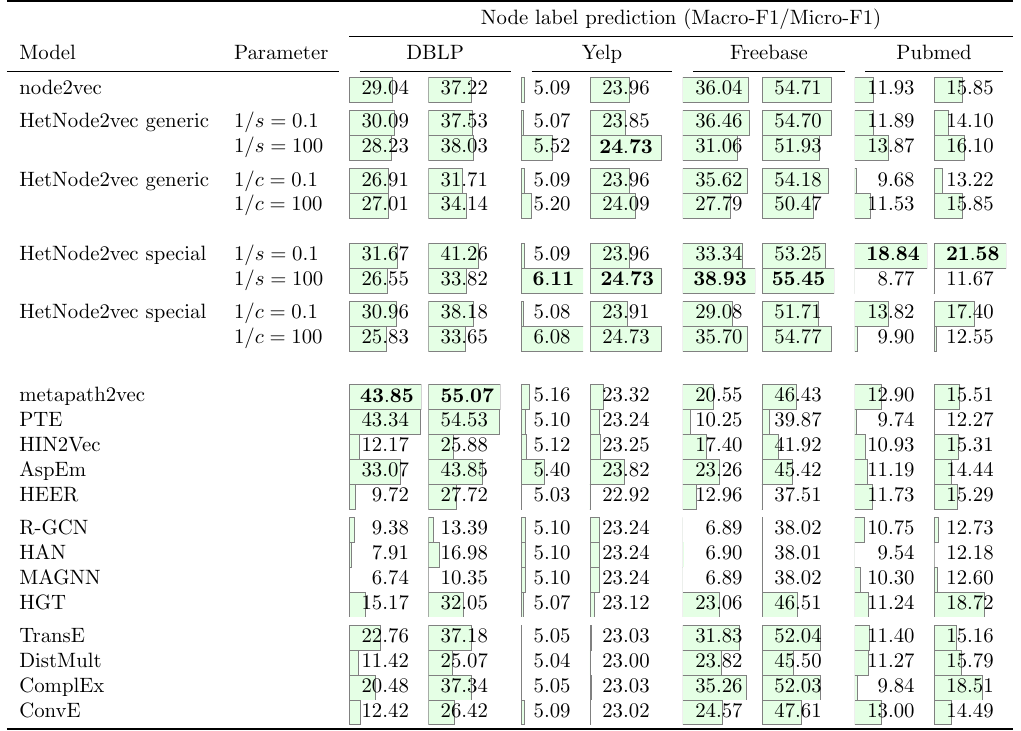}
\end{table}

Table~\ref{tab:link_prediction_results} in the Supplementary Material depicts the AUC and MRR measures computed to evaluate the link prediction tasks. 
In this task, \Het\ does not outperform the other competing methods but consistently ranks within the best five. 
This indicates that \Het\ can still produce valuable embeddings for link prediction, even if its primary strength lies in node-label prediction.

Nevertheless, prediction results also strongly depend on the supervised method applied to the computed embeddings.
Indeed, by repeating the same experiments using this time, a nonlinear classifier (a random forest), we significantly improved the edge prediction results using \Het edge embeddings (Table~\ref{tab:random_forest_results} in the Supplementary Material).
%, outperforming all the other methods on all the considered datasets 

 %%%%%%%%%%%%%%%%%%%%%%%%%%%%%%%%%%%%%%%%
\section{Discussion and Conclusions} \label{sec:discussion}

  The results obtained through the proposed \Het algorithm indicate the advantages of embedding techniques that account for node and edge heterogeneity in graphs.
  Methods like \ntv, originally conceived for homogeneous graphs, even if can be applied to heterogeneous networks, may fail to capture the diverse types of nodes and relationships, often leading to sub-optimal performance in real-world applications.
  On the other hand, many state-of-the-art heterogeneous graph representation learning methods depend on the specific characteristics of the input graph, and/or rely on large, not-scalable neural network architectures, or are trained to produce embeddings that are tailored to a specific predictive task.  
  By introducing type-aware second-order random walks, \Het is a flexible and scalable alternative that captures the topological graph structure and also reflects the semantic diversity present in the graph. The embeddings produced by \Het are task independent and may be used for several downstream tasks.
  Our experiments on multiple benchmark datasets demonstrate that our model enables the design of simple strategies to balance the exploration of graph heterogeneity.
  For example, datasets like Freebase and PubMed benefit from increasing the heterogeneity in the random walk, promoting the representation of the relationships between heterogeneous nodes to capture the enriched semantic of heterogeneous graphs.
  Conversely, for datasets where nodes are more evenly distributed, such as Yelp, promoting homogeneity in the walks can improve performance by focusing the walk on specific target node types. This can also be beneficial when we focus on underrepresented nodes or edges in the underlying heterogeneous graph.
  Indeed, our results suggest that in cases where certain node types dominate, focusing on special node types can yield significant improvements in the resulting representation.
  On the other hand, excessive focus on these special nodes can result in overly homogeneous embeddings that neglect the broader network structure, thereby limiting the representation ability of the methods.

  Our results show that by experimentally tuning the $s$ and $c$ parameters, we can significantly improve results with respect to homogeneous methods. Moreover, extensive experimental results outline that \Het\ is competitive with state-of-the-art methods for heterogeneous graphs and exhibits a temporal complexity of the same order of its homogeneous counterpart \ntv, which allows for scalable applications with large heterogeneous networks.

  Although the proposed method is able to match or outperform alternative approaches, its representation potential is far from being completely exploited and should be further explored as a future work. 
  
  Indeed, our future works will delve into a more detailed analysis of the relationship between the node and edge type distributions and the parametrization of the model, by introducing techniques to automatically learn, and dynamically update, the switching strategy.
  %This will enable the design of strategies that can optimally exploit the rich semantic information in the graph.
  
  \section*{Code and data availability}
\Het is implemented using the efficient RW-generation provided by the GRAPE library \cite{grape23}, which enables efficient first and \so RW generation through the usage of efficient and succinct data structures and an optimized Rust implementation with Python binders.
The \Het~code is available at \url{https://github.com/AnacletoLAB/hetnode2vec_ensmallen}.
The datasets and the benchmark pipeline used in the experiments follow the experimental set-up proposed  in~\cite{yang2020heterogeneous} and are available from  \url{https://github.com/yangji9181/HNE}.\\

\section*{Funding}
This paper is supported by FAIR (Future Artificial Intelligence Research) project, funded by the NextGenerationEU program within the PNRR-PE-AI scheme (M4C2, Investment 1.3, Line on Artificial Intelligence),
and by the  National Center for Gene Therapy and Drugs Based on RNA Technology—MUR (Project no. CN\_00000041) funded by NextGeneration EU program.

%%%%%%%%%%%%%%%%%%%%%

\bibliographystyle{plain}
%\bibliography{biblio}

\newpage
\appendix
\begin{center}
{\LARGE \bf Supplementary Information}  
\end{center}

%%%%%%%%%%%%%%%%%%%%%%%%%%%%%%%
\section{Impact of \ntv parameters on node embeddings}
\label{app:pq_effects}
\begin{figure}[htbp]
   \centering
   \begin{subfigure}[b]{\linewidth}
     \centering
     \includegraphics[width=0.2\linewidth]{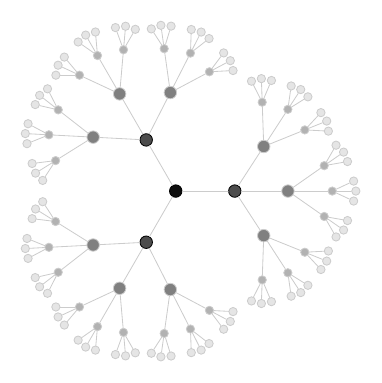}
     \hspace{20pt}
     \includegraphics[width=0.2\linewidth]{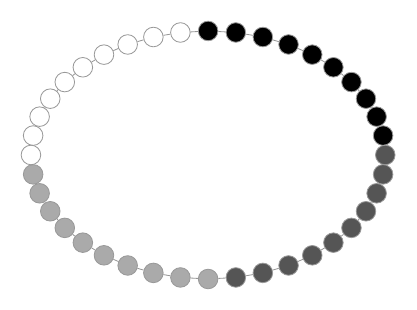}
     \caption{Graph topologies, the colors of the nodes represent their position in the graph.}
     \label{fig:topologies_p}
   \end{subfigure}
   \hfill
   \begin{subfigure}[b]{\linewidth}
     \centering
     \includegraphics[width=\linewidth]{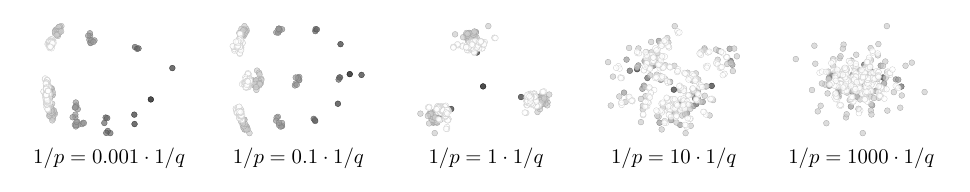}
     \caption{PCA projections of the \ntv embeddings of the tree using a target space of dimension 25.} 
     \label{fig:tree_size_25_p_PCA}
   \end{subfigure}
   \hfill
   \begin{subfigure}[b]{\linewidth}
     \centering
     \includegraphics[width=\linewidth]{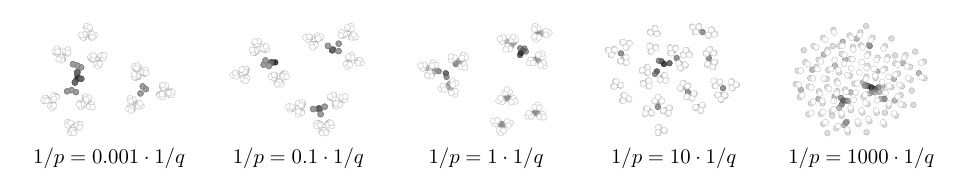}
     \caption{TSNE projections of the \ntv embeddings of the tree using a target space of dimension 25.}
     \label{fig:tree_size_25_p_TSNE}
   \end{subfigure}
   \hfill
   \begin{subfigure}[b]{\linewidth}
     \centering
     \includegraphics[width=\linewidth]{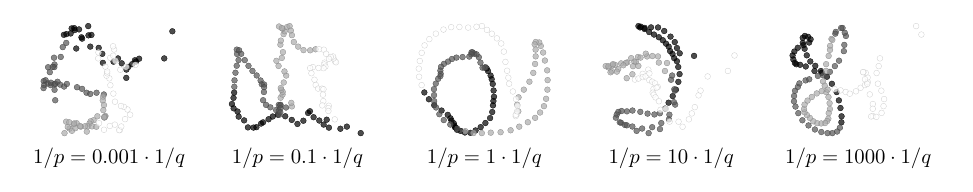}
     \caption{PCA projections of the \ntv embeddings of the cycle using a target space of dimension 5.}
     \label{fig:cycle_size_5_p_PCA}
   \end{subfigure}
   \hfill
   \begin{subfigure}[b]{\linewidth}
     \centering
     \includegraphics[width=\linewidth]{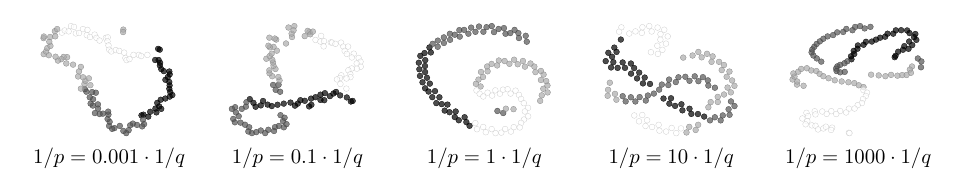}
     \caption{TSNE projections of the \ntv embeddings of the cycle using a target space of dimension 5.}
     \label{fig:cycle_size_5_p_TSNE}
   \end{subfigure}
   \caption{The figures depict the embeddings obtained for a 3-ary tree and a cycle using different values for the parameter $p$ and $q$. The resulting path were embedded using the Skipgram architecture into a space of dimension 25 for the tree and 5 for the cycle, respectively. Final bi-dimensional projections were obtained using their first two principal components and TSNE.}
   \label{fig:embedding_p}
      %\Description{Embeddings obtained using  node2vec for different hyperparameters configurations in two simple graph topologies: a 3-ary tree and a cycle.}
 \end{figure}

 Figure~\ref{fig:embedding_p} shows the embedding obtained by using different values for the inward and in-out parameters in two simple graphs: a 3-ary tree of depth five and a cycle of length 120. Graph topologies are outlined in Figure~\ref{fig:topologies_p}.
  In these examples, the graphs are assumed to be unweighted, while the color of a node represents its global position in the graph.
  Namely, in the tree, the color of a node indicates its depth, while in the cycle, it represents the region where the node is placed according to a cyclic visit.

  Notice that none of the graphs in the example contains a triangle (a complete subgraph of three vertices); therefore, the transition probabilities along the generation of a RW can take only the values $1/p$ or $1/q$.
  Therefore, the ratio $q/p$ fully characterizes the parameter space by quantifying the relation between the local/global type of the generated paths.    
   
Figures~\ref{fig:tree_size_25_p_PCA},\ref{fig:tree_size_25_p_TSNE},
\ref{fig:cycle_size_5_p_PCA} and \ref{fig:cycle_size_5_p_TSNE}  show the projections of the embeddings derived from \ntv for varying ratios of parameters $p$ and $q$.
  The corpus of the embedding contains, starting from each node, ten random walks of length 128; the paths that resulted from these random walks were embedded using the Skipgram method with a window size of five, and embedded into a space of dimensions 25 for the tree and 5 for for the cycle, respectively.
  These embeddings were then projected into a two-dimensional space using both their first two principal components (PCA) and TSNE.
  
  We notice that if the probability of return is smaller than that of exploration ($q/p<1$), the resulting embedding better captures the global structure of the input graph.
  Conversely, if the return probability is larger, the embedding depicts the local structure of the graph around each node.  

\newpage 
%%%%%%%%%%%%%%%%%%%%%%%%%%%%%%%%%%%%%%%%%%%%%%%%%%
\section{Impact of \Het parameters on the grid embedding}
\label{sec:appendix_grid_embedding}
 \begin{figure}[htbp]
   \centering
   % \begin{subfigure}[b]{\linewidth}
   %   \centering
   %   \includegraphics[width=0.3\linewidth]{grid_topology}
   %   \caption{Grid graph topology, the colors of a node represents its type.}
   %   \label{fig:tree_topology}
   % \end{subfigure}
   % \hfill
   % \begin{subfigure}[b]{\linewidth}
   %   \centering
   %        \includegraphics[width=\linewidth]{grid_size_10_p_PCA}
   %   \caption{PCA}
   %   \label{fig:grid_size_10_p_PCA}
   % \end{subfigure}
   % \hfill
   % \begin{subfigure}[b]{\linewidth}
   %   \centering
  \hspace{-50pt}\includegraphics[width=1\linewidth]{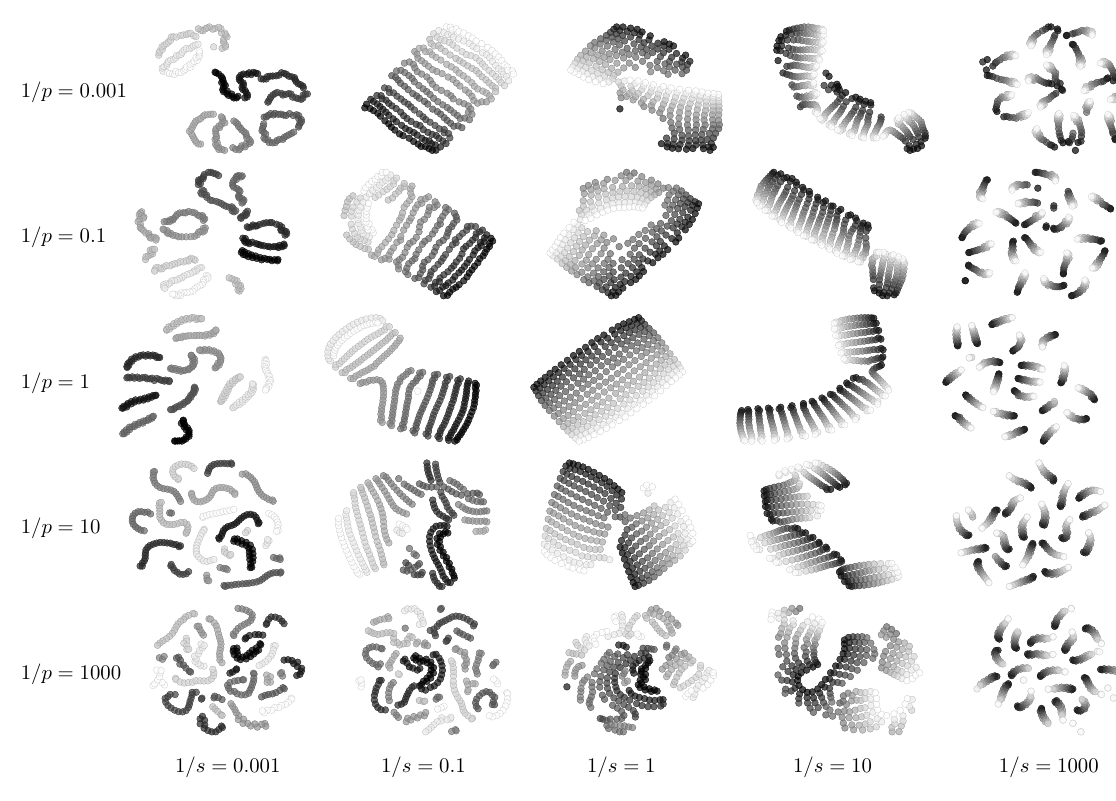}
     \caption{TSNE}
     \label{fig:grid_size_10_p_s_TSNE}
   % \end{subfigure}
   \caption{Embedding obtaining for a grid using different values for the parameter $p$, which regulates the exploration tendency of the generated random paths. Random paths were generated with node2vec algorithm using $q=1$ and a path length equal to 128.
     The resulting paths were embedded using Skipgram, with a window of size five, into a space of dimensions 10 and then projected using TSNE.}
     %\Description{Embeddings obtained using Het-node2vec for different hyperparameters values in the grid where types are defined by the grid rows.}
   \label{fig:grid_size_10}
 \end{figure}

%%%%%%%%%%%%%%%%%%%%%%%%%%%%%%%%%%%%%%%%%%%%%%%%%%%%%%%%%
\section{Equivalence of transition probabilities on special node-switching strategies}
\label{sec:special_node_switching_comparison}
In this section we show the equivalence of the normalized transition probabilities computed by \Het\ when we use the two functions $\beta^{(k)}_s$  and $\bar\beta^{(k)}_s$ introduced in Section 3.3 of the main manuscript.

Let $\beta^{(1)}_s$ and $\beta^{(2)}_s$ two special node-type switching strategies defined in Equation~7 and~8 of the main manuscript:
\begin{equation*}
      \beta^{(1)}_s(v,x) =
      \begin{cases}
        \frac{1}{s} & \text{ if } \phi(x) \in \Sigma_{\phi_\mathcal{S}}\\
        1 & \text{otherwise}.
      \end{cases}
      \quad\text{ and }\quad
            \beta^{(2)}_s(v,x) =
      \begin{cases}
        \frac{1}{s} & \text{ if } \phi(v) \notin \Sigma_{\phi_\mathcal{S}} \wedge \phi(x) \in \Sigma_{\phi_\mathcal{S}}\\
        1 & \text{otherwise}.
      \end{cases}
\end{equation*}

On the other hand, let $\bar\beta^{(1)}_s$ and $\bar\beta^{(2)}_s$  the implementations for these strategies defined in  Equation~10 %~XR \ref{eq:impl_special_node_beta1}
and~11:% XR \ref{eq:impl_special_node_beta2}:

\begin{equation*}
  \bar\beta^{(1)}_s(v,x) =
  \begin{cases}
    \frac{1}{s} & \text{ if } \phi(v) \in \Sigma_{\phi_\mathcal{S}}\; \underline\vee\; \phi(x) \in \Sigma_{\phi_\mathcal{S}}\\
    \frac{1}{s^2} & \text{ if } \phi(v) \in \Sigma_{\phi_\mathcal{S}} \wedge \phi(x) \in \Sigma_{\phi_\mathcal{S}}\\
    1 & \text{otherwise},
  \end{cases}
  \quad \text{and} \quad
      \bar\beta^{(2)}_s(v,x) =
      \begin{cases}
        \frac{1}{s} & \text{ if } \phi(v) \in \Sigma_{\phi_\mathcal{S}} \vee \phi(x) \in \Sigma_{\phi_\mathcal{S}}\\
        1 & \text{otherwise}.
      \end{cases} 
    \end{equation*}

    We first notice that for both strategies $\beta^{(k)}_s,\, k\in\{1,2\}$ it holds that
    \begin{itemize}
    \item[(1)] if $v$ is a non-special node, then $\bar\beta^{(k)}_s(v,x)=\beta^{(k)}_s(v,x)$ and,
    \item[(2)] if $v$ is a special node, then $\bar\beta^{(k)}_s(v,x)=\tfrac1s\beta^{(k)}_s(v,x)$. 
    \end{itemize}

    In the following, we show that transition probabilities induced by the strategies $\beta^{(1)}_s$ and $\beta^{(2)}_s$ are equal to those defined by their implementations   $\bar\beta^{(1)}_s$ and $\bar\beta^{(2)}_s$ respectively.

    Let $v$ be the node currently visited by a \RW coming from a node $r$ and generated according to the strategy $\bar\beta^{(k)}_s$.
    We observe that there are two possible cases for node $v$, i.e. $v$ maybe or maybe not a special node, and the functions $\beta^{(k)}_s$ completely define the possible switching to a node $x$ that can be special or non special. Hence, the normalized transition probability from node $v$ to its neighbours $x\in\mathcal{N}(v)$ includes two possible cases:
    
    \begin{itemize}
    \item[Case 1.] If $v$ is a non-special node then 
\begin{align*}    
  P\Bigl(X_{t+1}=x, E_{t+1} = e_{vx} | X_t=v, X_{t-1}=r, E_{t} = e_{rv} \Bigr)
  &=\frac{\bar\beta^{(k)}_s(v,x)\cdot \gamma_c(e_{rv},e_{vx})\cdot \alpha_{pq}\cdot  w_{e_{vx}}}{\textstyle \sum\limits_{vz\in E}\bar\beta^{(k)}_s(v,z)\cdot \gamma_c(e_{rv},e_{vz})\cdot \alpha_{pq}\cdot  w_{e_{vz}}} \\
  &=\frac{\beta^{(k)}_s(v,x)\cdot \gamma_c(e_{rv},e_{vx})\cdot \alpha_{pq}\cdot  w_{e_{vx}}}{\textstyle \sum\limits_{vz\in E}\beta^{(k)}_s(v,z)\cdot \gamma_c(e_{rv},e_{vz})\cdot \alpha_{pq}\cdot  w_{e_{vz}}}.
\end{align*}

\item[Case 2.] If $v$ is a special node then
\begin{align*}    
  P\Bigl(X_{t+1}=x, E_{t+1} = e_{vx} | X_t=v, X_{t-1}=r, E_{t} = e_{rv} \Bigr)
  &=\frac{\bar\beta^{(k)}_s(v,x)\cdot \gamma_c(e_{rv},e_{vx})\cdot \alpha_{pq}\cdot  w_{e_{vx}}}{\textstyle \sum\limits_{vz\in E}\bar\beta^{(k)}_s(v,z)\cdot \gamma_c(e_{rv},e_{vz})\cdot \alpha_{pq}\cdot  w_{e_{vz}}}\\
  &=\frac{\tfrac{1}{s}\beta^{(k)}_s(v,x)\cdot \gamma_c(e_{rv},e_{vx})\cdot \alpha_{pq}\cdot  w_{e_{vx}}}{\textstyle \sum\limits_{vz\in E}\tfrac{1}{s}\beta^{(k)}_s(v,z)\cdot \gamma_c(e_{rv},e_{vz})\cdot \alpha_{pq}\cdot  w_{e_{vz}}}.\\
  &=\frac{\beta^{(k)}_s(v,x)\cdot \gamma_c(e_{rv},e_{vx})\cdot \alpha_{pq}\cdot  w_{e_{vx}}}{\textstyle \sum\limits_{vz\in E}\beta^{(k)}_s(v,z)\cdot \gamma_c(e_{rv},e_{vz})\cdot \alpha_{pq}\cdot  w_{e_{vz}}}.\\
\end{align*}
\end{itemize}

In both cases the strategies $\beta^{(k)}_s$ and $\bar\beta^{(k)}_s$ define the same transition probabilities, hence $\beta^{(k)}_s$ and $\bar\beta^{(k)}_s$ model the same type-aware switching strategy. 

\begin{table}[htbp]
    \caption{Performance metrics comparison between the node-switching strategies for the link prediction and edge prediction tasks on a benchmark dataset (Pubmed). The dataset is described in Section~\ref{app:datasets} while the experimental setting for the supervised tasks is described in Section~4.2}%\ref{sub:analysis-switch-real}.}
    \label{tab:strategies_comparison}
    \centering
\includegraphics[width=.8\linewidth]{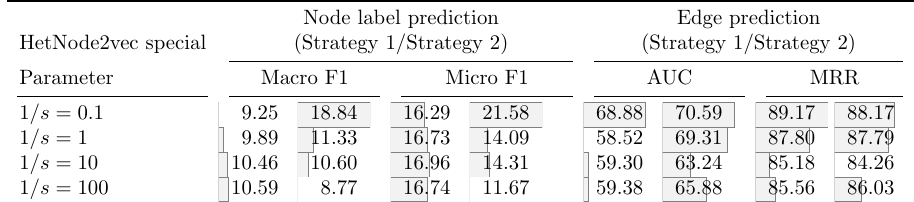}    
\end{table}

 \newpage

%%%%%%%%%%%%%%%%%%%%%%%%%%%%%%%%%%%%%%%%%%%%%%%%%%%%%%%%%
\section{Supplementary experimental analysis of the switching parameters of \Het.}
\label{supp:switching-exp}
\begin{figure}[htbp]
   \centering
   \begin{subfigure}[b]{\linewidth}
     \centering
     \scalebox{-1}[1]{% horizontal flip
       \includegraphics[width=0.4\linewidth]{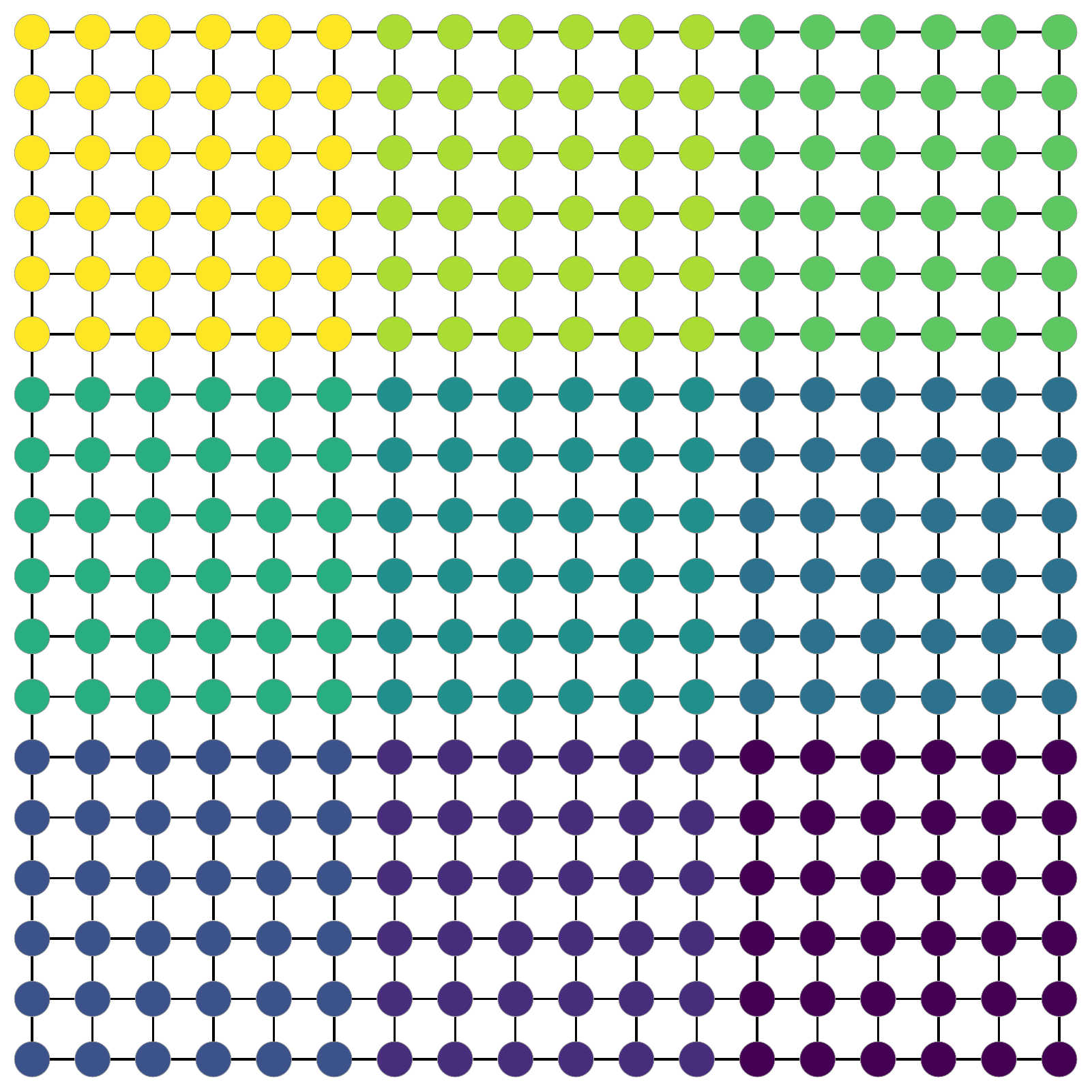}
       }
     \caption{A grid structure with groups; the graph is generated by considering a grid partitioned into different regions composed of intervals of contiguous rows and columns defining a node type.}
     \label{fig:grid_wiht_groups_topology}
   \end{subfigure}
   \hfill
   \begin{subfigure}[b]{\textwidth}
     \includegraphics[width=1.1\textwidth]{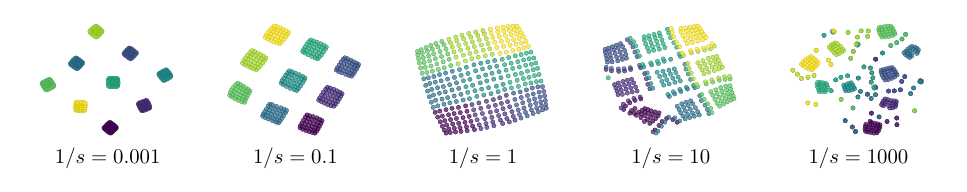}
     \caption{\Het~embeddings of the clustered grid for different node type switching values $s$ and $p=q=1$.}
   \label{fig:clusterd_grid_size_10_p_TSNE}
   \end{subfigure}
   \begin{subfigure}[b]{\textwidth}
     \includegraphics[width=1.1\textwidth]{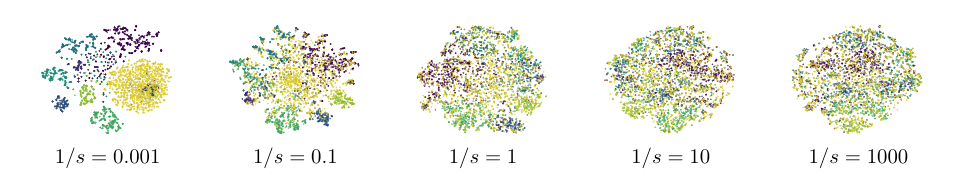}
     \caption{\Het~embeddings of the Cora graphs for different node type switching values $s$ and $p=q=1$.}
   \label{fig:cora_size_10_p_TSNE}
   \end{subfigure}
   \caption{Embedding obtained by using different values for the node switching parameter $s$ ($p=1$, $q=1$).
   Using this \Het setting, embeddings are computed by generating 10 RWs of length 128 from each node. These paths are input to a Skipgram architecture, with a window size equal to 5, to generate an embedding of the nodes into a 10D space. The final 2D representation is obtained by t-SNE.
   (a) A grid topology divided into nine node types according to their region.
   (b) Embeddings obtained from the clustered grid.
   (c) Embeddings obtained from the Cora network.}
  \label{fig:clustered-grid}
     %\Description{Embeddings obtained using Het-node2vec for different hyperparameters values in of the grid where types are defined as square sub-grids.}
\end{figure}

% The Supplementary Fig.~\ref{fig:clustered-grid} shows the behaviour of \Het\ by varying its $s$ switching parameter using both synthetic and real-world data sets.
Figure~\ref{fig:clustered-grid} shows how \Het behaves on synthetic and real-world data sets when changing the $s$ switching parameter.

 %%%%%%%%%%%%%%%%%%%%%%%%%%%%%%%%%%%%%%%%%%%

%%%%%%%%%%%%%%%%%%%%%%%%%%%%%%%%%%%%%%%%%%%%%%%%%%%%%%%%%
\section{Datasets}
\label{app:datasets}
  \paragraph{The Freebase network}  is constructed based on the collaborative knowledge base Freebase\footnote{\url{http://www.freebase.com}}, containing relations between books, films, music, sports, people, locations, organizations, and businesses. Each node possesses a unique type and no features, meaning that the nodes are not attributed, while each edge is defined by a unique type characterized by the types of its endpoint nodes.

\begin{figure}[tb]
    \centering
    \begin{subfigure}{0.48\textwidth}
      \centering
      \includegraphics[width=.85\linewidth]{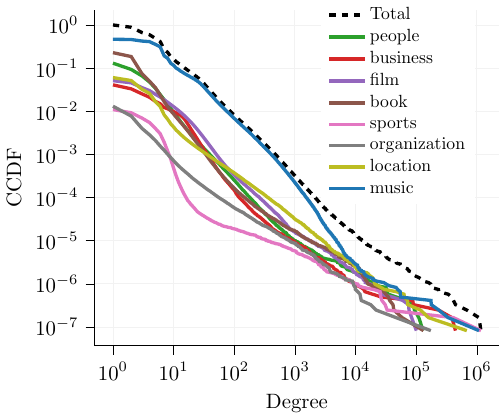}
      \caption{Freebase}
      \label{fig:ccdf_freebase}
      \end{subfigure}
    \begin{subfigure}{0.48\textwidth}
      \centering      
      \includegraphics[width=.85\linewidth]{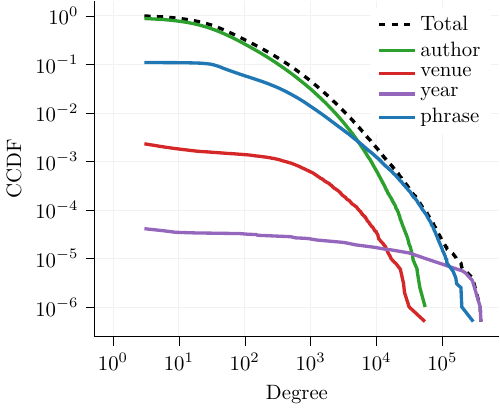}
      \caption{DBLP}
      \label{fig:ccdf_DBLP}
      \end{subfigure}
    \begin{subfigure}{0.48\textwidth}
      \centering      
      \includegraphics[width=.85\linewidth]{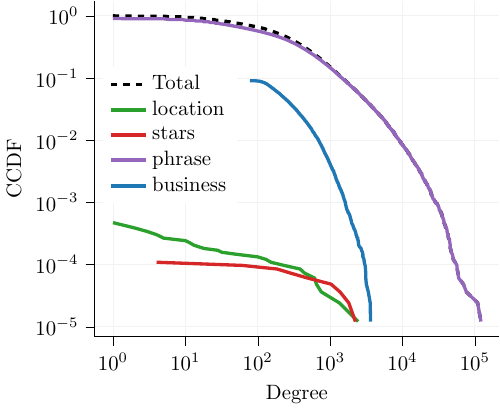}
      \caption{Yelp}
      \label{fig:ccdf_yelp}
      \end{subfigure}
    \begin{subfigure}{0.48\textwidth}
      \centering      
      \includegraphics[width=.85\linewidth]{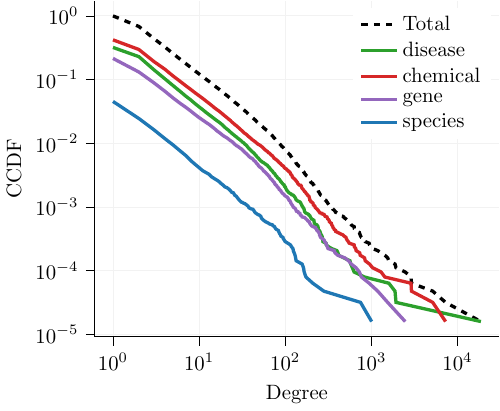}
      \caption{PubMed}
      \label{fig:ccdf_pubmed}
      \end{subfigure}
  \caption{Complementary Cumulative Distribution Function of degrees with respect to their node types.}
  \label{fig:ccdf}
  %\Description{Plots of the Complementary Cumulative Distributions Functions of the degrees for each node type of all benchmark datasets. }
\end{figure}

Freebase is the largest of the graphs concerning node-cardinality and exhibits the highest number of types for nodes and edges, comprising eight node types connected by 36 edge types. Both node and edge types have a highly unbalanced distribution, where two node/edge types are significantly more prevalent compared to others (see Table~\ref{tab:node_type_distribution} 
and Figure~\ref{fig:edge_type_distribution}
).
As shown in the CCDF in Figure~\ref{fig:ccdf_freebase}
, the degree distribution of the nodes shows a long tail across all types of nodes, with the \texttt{music} nodes exhibiting the highest mean degree (mostly connected to other nodes), while the (target) \texttt{book} nodes have the lowest mean degree.

\paragraph{The DBLP network} is derived from the well known dataset DBLP\footnote{\url{https://dblp.org}}, and collects bibliographical information related to computer science publications.  
 The creators of the dataset~[47] %\cite{yang2020heterogeneous}
  built on DBLP dataset to construct an attributed multi-graph where node types are authors, phrases, venues, and years; and where node attributes are characterized by a 300-dimensional feature vector.
 Phrases nodes represent ``quality phrases'' extracted semi-automatically from each paper using the AutoPhrase algorithm  %\cite{shang2018automated}
 , and followed by human curation.
 Each phrase node is connected to the authors/venue/year nodes associated with the paper from which the phrase was sourced.
 For the construction of attributes for \texttt{phrases} and \texttt{paper} nodes, authors aggregated the word2vec representations of their constituent words.
 For \texttt{author}, \texttt{venue}, \texttt{year} nodes, attributes are formed by the aggregation of the feature vectors derived from their related papers (for instance, those published by an author, in a particular venue, or within a certain year).
Additionally, as a result of a web mining process, a small group of authors have been categorized into 12 research groups spanning four research areas and used for the node-label prediction task.

  DBLP ranks second in terms of node cardinality, but it exhibits the largest number of edges and therefore the highest values for the node degrees.
  
\paragraph{The Yelp network} represents relationships between reviews (phrases), businesses, locations, and stars (rating) nodes extracted from the Yelp dataset\footnote{\url{https://www.yelp.com/dataset/challenge}}.
  As shown in Table~\ref{tab:node_type_distribution}, \texttt{phrase} nodes form most of the graph, covering close to 91\% of the overall nodes. Nodes of type $business$ follow next accounting for almost 9\% of the total nodes, while the remaining two types, \texttt{location} (25 nodes) and \texttt{stars} (9 nodes),  represent a negligible portion of the network and could be viewed as hub nodes.

\paragraph{The PubMed network} is constructed starting from the PubMed database~\footnote{\url{https://www.ncbi.nlm.nih.gov/pubmed/}}, and contain nodes representing genes, diseases, chemicals, and species.
  All nodes are extracted from PubMed papers through the  AutoPhrase algorithm%~\cite{shang2018automated}
  , nodes are typed using  bioNER%~\cite{shang-etal-2018-learning}
  , and filtered by human experts.
  All the nodes in the networks are attributed.
  To construct the node features, PubMed papers were represented by 200 dimensional vectors via word2vec, and node vectors were subsequently obtained through an aggregation process.
  
  Among the networks, PubMed is the smallest one with all node types exhibiting a long-tail distribution on their degrees (see Figure~\ref{fig:ccdf_pubmed}).
  Compared to other networks, PubMed displays a lower level of unbalancing of nodes types, edge types, and (disease) label distribution,  as shown in 
  Table~\ref{tab:node_type_distribution}
  and Figures~\ref{fig:edge_type_distribution},\ref{fig:label_distribution}.

%%%%%%%%%%%%%%%%%%%%
\section{Prediction tasks on real-world data sets}
\label{suppl:pred-tasks}
\paragraph{Predictive tasks.}
With the Freebase dataset,
node-label classification is applied to nodes of type $book$, the second most common node type covering 22.8\% of the overall nodes, surpassed only by $music$ nodes (46.6\% --  Table~\ref{tab:node_type_distribution}).
A significant portion of book nodes has been labeled into eight unbalanced literature genres, resulting in an unbalanced multi-class classification problem (Figure~\ref{fig:label_distribution_freebase}).
The edge prediction task is performed on the \texttt{book-book} edge type, accounting for 5.95\% of the edges. It is the second most common edge type after edges connecting two \texttt{music} nodes (\texttt{music-music}), which make up 61.77\% of the edges (Figure~\ref{fig:edge_type_distribution_freebase}).

With DBLP, node-type prediction task is performed with {\em author} nodes.
The authors have been categorized into 12 research groups spanning four research areas and used for the node-label prediction task.
In this graph, \texttt{author} and \texttt{phrases} nodes represent almost the entire graph, accounting for around 96\% of both nodes and edges.
The edge-prediction task is focused on the two \texttt{author-author} edge types,  i.e. the relations of type \texttt{co-author} and \texttt{cite}, which account for about the 59\% of all edges in the graph (Figure~\ref{fig:edge_type_distribution_freebase}).

With Yelp, the \texttt{business} node has been selected for the node-label prediction task. Business nodes account for about 9\% of the overall nodes and are labeled with one or multiple categories among sixteen potential ones.  These categories have an uneven distribution, which results in an unbalanced multi-label prediction (Figure~\ref{fig:label_distribution_yelp}). 
For edge prediction the \texttt{business-phrase} has been selected,
thus resulting in a prediction task involving almost the 9\% of the overall Yelp edges (Figure \ref{fig:edge_type_distribution_yelp}).

Node-label prediction task in PubMed is applied to \texttt{disease} nodes, which were labeled into eight categories, thus resulting in a multi-class classification problem.
Disease nodes make up approximately 32\% of the total nodes, representing the second most common type. \texttt{Disease-disease} edges, on which edge prediction task is focused, constitute about 14\% of the network edges (Figure \ref{fig:edge_type_distribution_pubmed}).

\begin{table}[htb]
            \caption{Node type distribution and basic degree statistics in the four heterogeneous network datasets. 
            Proportion refers to the ratio of a given type of node with respect to the total number of nodes.
            The target node types used for the node type prediction task are highlighted in bold.
      }
      \label{tab:node_type_distribution}
    \centering
%    \scalebox{.9}{
      \begin{tabular}{l lr rrrr} 
      \toprule
        Graph & Node type                                  & Proportion &\multicolumn{3}{c}{Degree}\\
        \cmidrule(l){4-6} 
              &                                       &                                                       & mean                       & min                   & max                        \\
      \midrule
     Freebase
               & music                                 & \blackwhitebar{0.466}                     & 14.52                      & 1                     & 1,086,802                  \\
               & \cellcolor{gray!15}\textbf{book}      & \cellcolor{gray!15}\blackwhitebar{0.228}  & \cellcolor{gray!15} 3.73   & \cellcolor{gray!15} 1 & \cellcolor{gray!15}131,957 \\
               & people                                & \blackwhitebar{0.130}                     & 5.17                       & 1                     & 130,116                    \\
               & location                              & \blackwhitebar{0.060}                     & 9.35                       & 1                     & 684,726                    \\
               & film                                  & \blackwhitebar{0.051}                     & 11.36                      & 1                     & 100,825                    \\
               & business                             & \blackwhitebar{0.041}                     & 11.15                      & 1                     & 445,716                    \\
               & organization                          & \blackwhitebar{0.013}                     & 7.85                       & 1                     & 174,646                    \\
               & sports                                & \blackwhitebar{0.011}                     & 19.289                     & 1                     & 1,170,520                  \\
     \midrule
        DBLP
               & \cellcolor{gray!15}\textbf{author}    & \cellcolor{gray!15}\blackwhitebar{0.8880} & \cellcolor{gray!15} 188.68 & \cellcolor{gray!15} 1 & \cellcolor{gray!15} 71,861 \\
               & phrase                                & \blackwhitebar{0.1094}                    & 739.55                     & 1                     & 294,453                    \\
               & venue                                 & \blackwhitebar{0.0026}                    & 981.97                     & 1                     & 54,396                     \\
               & year                                  & \blackwhitebar{0.00004}                    & 58742.77                   & 1                     & 385,014                    \\
     \midrule
        Yelp
               & phrase                                & \blackwhitebar{0.9088}                    & 761.57                     & 1                     & 121,333                    \\ 
               & \cellcolor{gray!15}\textbf{business} & \cellcolor{gray!15}\blackwhitebar{0.0906} & \cellcolor{gray!15}357.14  & \cellcolor{gray!15}78 & \cellcolor{gray!15} 3,625  \\
               & location                              & \blackwhitebar{0.0005}                    & 191.64                     & 1                     & 2,443                      \\ 
               & stars                                 & \blackwhitebar{0.0001}                    & 830.44                     & 4                     & 2,239                      \\
        \midrule
        Pubmed
               & chemical                              & \blackwhitebar{0.420}                     & 8.04                       & 1                     & 7,272                      \\ 
               & \cellcolor{gray!15}\textbf{disease}   & \cellcolor{gray!15}\blackwhitebar{0.319}  & \cellcolor{gray!15}7.48    & \cellcolor{gray!15}1  & \cellcolor{gray!15}18,714  \\ 
               & gene                                  & \blackwhitebar{0.215}                     & 6.83                       & 1                     & 2,474                      \\    
               & species                               & \blackwhitebar{0.045}                     & 5.69                       & 1                     & 1,008  \\
      \bottomrule
      \end{tabular}
%      }
  \end{table}

%%%%%%%%%%%%%%%%%%%%%%%%%%%%%%%%%%%%%%%%%%%%%%%%%%%%%%%%%%%%%%%
%%%%%%%%%%%%%%%%%%%%%%%%%%%%%%%%%%%%%%
\begin{figure}[htb]
  \centering
  \begin{minipage}{.45\linewidth}
    \begin{subfigure}{\linewidth}
      \centering
    \resizebox{.8\linewidth}{!}{
    \begin{tabular}[t]{l r }
      \\  \multicolumn{2}{c}{edge-type distribution}
      \\
        \cmidrule(lr){1-2}  
 music-and-music               & \blackwhitebar[0.6177]{0.6177} \\
\cellcolor{gray!15}book-and-book  & \cellcolor{gray!15}\blackwhitebar[0.6177]{0.0595} \\
business-about-music          & \blackwhitebar[0.6177]{0.0436} \\
location-and-location         & \blackwhitebar[0.6177]{0.0348} \\
film-and-film                 & \blackwhitebar[0.6177]{0.0347} \\
people-and-people             & \blackwhitebar[0.6177]{0.0213} \\
people-to-sports              & \blackwhitebar[0.6177]{0.0197} \\
people-to-book                & \blackwhitebar[0.6177]{0.0191} \\
people-to-music               & \blackwhitebar[0.6177]{0.0187} \\
people-on-location            & \blackwhitebar[0.6177]{0.0134} \\
business-and-business         & \blackwhitebar[0.6177]{0.0132} \\
people-to-film                & \blackwhitebar[0.6177]{0.0114} \\
music-in-film                 & \blackwhitebar[0.6177]{0.0105} \\
book-on-location              & \blackwhitebar[0.6177]{0.0075} \\
music-in-book                 & \blackwhitebar[0.6177]{0.0075} \\
book-to-film                  & \blackwhitebar[0.6177]{0.006}  \\
music-on-location             & \blackwhitebar[0.6177]{0.0057} \\
location-in-film            & \blackwhitebar[0.6177]{0.0055} \\
people-in-business            & \blackwhitebar[0.6177]{0.0051} \\
sports-and-sports             & \blackwhitebar[0.6177]{0.0049} \\
sports-in-film                & \blackwhitebar[0.6177]{0.0046} \\
business-about-film           & \blackwhitebar[0.6177]{0.0043} \\
business-on-location          & \blackwhitebar[0.6177]{0.0039} \\
organization-to-music         & \blackwhitebar[0.6177]{0.003}  \\
business-about-book           & \blackwhitebar[0.6177]{0.003}  \\
book-about-organ              & \blackwhitebar[0.6177]{0.0029} \\
organization-on-location      & \blackwhitebar[0.6177]{0.0025} \\
organization-for-business     & \blackwhitebar[0.6177]{0.0025} \\
sports-on-location            & \blackwhitebar[0.6177]{0.0023} \\
organization-in-film          & \blackwhitebar[0.6177]{0.0023} \\
organization-and-organization & \blackwhitebar[0.6177]{0.0022} \\
people-in-organization        & \blackwhitebar[0.6177]{0.002}  \\
music-for-sports              & \blackwhitebar[0.6177]{0.0019} \\
book-on-sports                & \blackwhitebar[0.6177]{0.0017} \\
business-about-sport          & \blackwhitebar[0.6177]{0.0007} \\
organization-to-sport         & \blackwhitebar[0.6177]{0.0006} 
  \end{tabular}
}
\caption{Freebase}
\label{fig:edge_type_distribution_freebase}
\end{subfigure}
\end{minipage}
\begin{minipage}{.45\linewidth}
  \begin{subfigure}{\linewidth}
  \centering
 \resizebox{.8\linewidth}{!}{
\begin{tabular}[t]{ll r }
  \\
  \multicolumn{3}{c}{edge-type distribution}                 \\
  \midrule
  \rowcolor{gray!15}
 author-author & : cite       & \blackwhitebar[0.555]{0.555} \\
 phrase-phrase & : co-occur   & \blackwhitebar[0.555]{0.273} \\
 author-phrase & : study      & \blackwhitebar[0.555]{0.092} \\
  \rowcolor{gray!15}
 author-author & : co-author  & \blackwhitebar[0.555]{0.041} \\
 author-venue  & : publish-in & \blackwhitebar[0.555]{0.020} \\
 author-year   & : active-in  & \blackwhitebar[0.555]{0.019} \\

\end{tabular}
}
\caption{DLBP}
\label{fig:edge_type_distribution_dblp}

  \end{subfigure}
  \begin{subfigure}{\linewidth}
    \centering
    \resizebox{0.8\linewidth}{!}{
 \begin{tabular}[t]{l r}
   \\
  \multicolumn{2}{c}{edge-type distribution}  \\
  \midrule
  business-in-location  & \blackwhitebar[0.9197]{0.0003}  \\
  business-rate-stars    & \blackwhitebar[0.9197]{0.0003}  \\
  \rowcolor{gray!15}
  business-described with-phrase    & \blackwhitebar[0.9197]{0.0888}  \\
 phrase-context-phrase      & \blackwhitebar[0.9197]{0.9107}  \\
\end{tabular}
}
\caption{Yelp}
\label{fig:edge_type_distribution_yelp}
\end{subfigure}

\begin{subfigure}{\linewidth}
  \centering
   \resizebox{0.8\linewidth}{!}{
\begin{tabular}[t]{l r}
  \\
  \multicolumn{2}{c}{edge-type distribution}  \\
  \midrule
 chemical-and-chemical  & \blackwhitebar[0.263]{0.263}  \\
 chemical-in-disease    & \blackwhitebar[0.263]{0.217}  \\
\cellcolor{gray!15}disease-and-disease    &\cellcolor{gray!15} \blackwhitebar[0.263]{0.144}  \\
 chemical-in-gene      & \blackwhitebar[0.263]{0.132}  \\

 gene-causing-diserse   & \blackwhitebar[0.263]{0.111}  \\

 gene-and-gene          & \blackwhitebar[0.263]{0.068}  \\

 chemical-in-species    & \blackwhitebar[0.263]{0.027}  \\

 species-with-disease   & \blackwhitebar[0.263]{0.022}  \\

 species-with-gene      & \blackwhitebar[0.263]{0.013}  \\

 species-and-species    & \blackwhitebar[0.263]{0.003}  \\

\end{tabular}
}    
\caption{Pubmed}
\label{fig:edge_type_distribution_pubmed}
\end{subfigure}
\end{minipage}
\caption{Edge-type distribution of the training set for the benchmark datasets. 
The target edge type used for the link prediction task in each dataset is highlighted in gray.}
\label{fig:edge_type_distribution}
%\Description{Tables showing the type distribution on the benchmark datasets.}
\end{figure}

%%%%%%%%%%%%%%%%%%%%%%%%%%%%%%%%%%%%%%%%%%%%%
%%%%%%%%%%%%%%%%%%%%%%%%%%%%%%%%%%%%%%%%%%%%%%%%

%%%%%%%%%%%%%%%%%%%%%%%%%%%%%%%%%%%%%%%%%%%%%%%%%%%%%%%%%%%
%%%%%%%%%%%%%%%%%%%%%%%%%%%%%%%%%%%%%%
\begin{figure}[htb]
  \centering
    \begin{subfigure}{.45\linewidth}
     \centering
      \small
%   \resizebox{.9}{!}{
   \scalebox{.9}{
    \begin{tabular}[t]{p{4cm} r} 
  %\toprule
      \multicolumn{2}{c}{label distribution for node type \textbf{book} } \\
  \cmidrule(lr){1-2}  
Book        & \blackwhitebar[0.382]{0.085} \\
Film        & \blackwhitebar[0.382]{0.382} \\
Music       & \blackwhitebar[0.382]{0.240} \\
Sports      & \blackwhitebar[0.382]{0.024} \\
People      & \blackwhitebar[0.382]{0.145} \\
Location    & \blackwhitebar[0.382]{0.085} \\
Organization& \blackwhitebar[0.382]{0.047} \\
Business    & \blackwhitebar[0.382]{0.012}
    \end{tabular}
  }
\vspace{50pt}    
\caption{Freebase}
\label{fig:label_distribution_freebase}
\end{subfigure}
  \begin{subfigure}{.45\linewidth}
    \centering
    \small
   \scalebox{.9}{
    % \resizebox{.9\linewidth}{!}{
     \begin{tabular}[t]{p{4cm} r} 
       \multicolumn{2}{c}{label distribution for node type \textbf{author}} \\
       \cmidrule(lr){1-2} 
Class 1 & \blackwhitebar[0.159]{0.167} \\
Class 2  & \blackwhitebar[0.159]{0.039} \\
Class 3  & \blackwhitebar[0.159]{0.068} \\
Class 4  & \blackwhitebar[0.159]{0.029} \\
Class 5  & \blackwhitebar[0.159]{0.159} \\
Class 6  & \blackwhitebar[0.159]{0.037} \\
Class 7  & \blackwhitebar[0.159]{0.052} \\
Class 8  & \blackwhitebar[0.159]{0.086} \\
Class 9  & \blackwhitebar[0.159]{0.128} \\
Class 10 & \blackwhitebar[0.159]{0.023} \\
Class 11 & \blackwhitebar[0.159]{0.149} \\
Class 12 & \blackwhitebar[0.159]{0.040} \\
Class 13 & \blackwhitebar[0.159]{0.024} \\
\end{tabular}
}
\caption{DLBP}
\label{fig:label_distribution_dblp}
\end{subfigure}

 \vspace{10pt}
  \begin{subfigure}{.45\linewidth}
    \centering
    \small
%    \resizebox{.9\linewidth}{!}{
   \scalebox{.9}{
      \begin{tabular}[t]{p{4cm} r}
  \multicolumn{2}{c}{label distribution for node type \textbf{business}}\\
        \midrule
Shopping                    & \blackwhitebar[0.295]{0.043} \\
Event Planning \& Services  & \blackwhitebar[0.295]{0.032} \\        
Automotive                  & \blackwhitebar[0.295]{0.035} \\          
Italian                     & \blackwhitebar[0.295]{0.029} \\
Beauty \& Spas              & \blackwhitebar[0.295]{0.055} \\
Pizza                       & \blackwhitebar[0.295]{0.032} \\
Sandwiches                  & \blackwhitebar[0.295]{0.039} \\ 
Food                        & \blackwhitebar[0.295]{0.085} \\
Bars                        & \blackwhitebar[0.295]{0.071} \\
Breakfast \& Brunch         & \blackwhitebar[0.295]{0.044} \\ 
Restaurants                 & \blackwhitebar[0.295]{0.295} \\ 
American (Traditional)      & \blackwhitebar[0.295]{0.050} \\ 
Nightlife                   & \blackwhitebar[0.295]{0.074} \\ 
Burgers                     & \blackwhitebar[0.295]{0.024} \\ 
Mexican                     & \blackwhitebar[0.295]{0.043} \\ 
American (New)              & \blackwhitebar[0.295]{0.046} \\ 
\end{tabular}
}
\caption{Yelp}
\label{fig:label_distribution_yelp}
\end{subfigure}
%\hfill
\begin{subfigure}{.45\linewidth}
  \centering
  \small
   \scalebox{.9}{
%   \resizebox{.9\linewidth}{!}{
     \begin{tabular}[t]{p{4cm} r} 
       \multicolumn{2}{c}{label distribution for node type \textbf{disease}}\\
 \cmidrule(l){1-2} 
Cardiovascular disease & \blackwhitebar[0.216]{0.152} \\ 
Glandular disease      & \blackwhitebar[0.216]{0.115} \\ 
Nervous disorder       & \blackwhitebar[0.216]{0.105} \\    
Communicable disease   & \blackwhitebar[0.216]{0.095} \\ 
Inflammatory disease   & \blackwhitebar[0.216]{0.216} \\
Pycnosis               & \blackwhitebar[0.216]{0.141} \\
Skin disease           & \blackwhitebar[0.216]{0.083} \\
Cancer                 & \blackwhitebar[0.216]{0.093} \\
\end{tabular}
}
\vspace{81pt}
 \caption{Pubmed} 
\label{fig:label_distribution_pubmed}
\end{subfigure}

\caption{Node-label distribution of target node types in the four heterogeneous network dataset benchmarks.}
\label{fig:label_distribution}
\end{figure}

\clearpage

%%%%%%%%%%%%%%%%%%%%%%%%%%%%%%%

\section{\Het\ results for the edge prediction tasks on real-world data sets}
\label{sec:edge_prediction}
Table~\ref{tab:link_prediction_results} summarizes the results of the comparison of \Het\ with other state-of-the-art methods for edge prediction on real-world benchmark data sets.
Table~\ref{tab:random_forest_results} outlines the results achieved by \Het\ when their node and edge type-aware embeddings are used to train a Random Forest for node label and edge prediction problems.

%%%%%%%%%%%%%%%%%%%%%%%%%%%%
    \begin{table}[htb]
          \caption{Performance metric for the link prediction task on the benchmark graphs. In the table, for both the \Het settings, we report only the values obtained by the two extreme values of the node-type switching parameter $1/s = {0.1, 100}$. To avoid confounding effects, when using a (generic or special) node-type switching strategy $1/s >0$, the (generic or special) edge-type switching is disabled ($1/c=0$), and viceversa. When using the special edge-type switching strategy, the special edge-type is the one targeted by link prediction.
      }
  \label{tab:link_prediction_results}
      \centering
      \includegraphics[width=\textwidth]{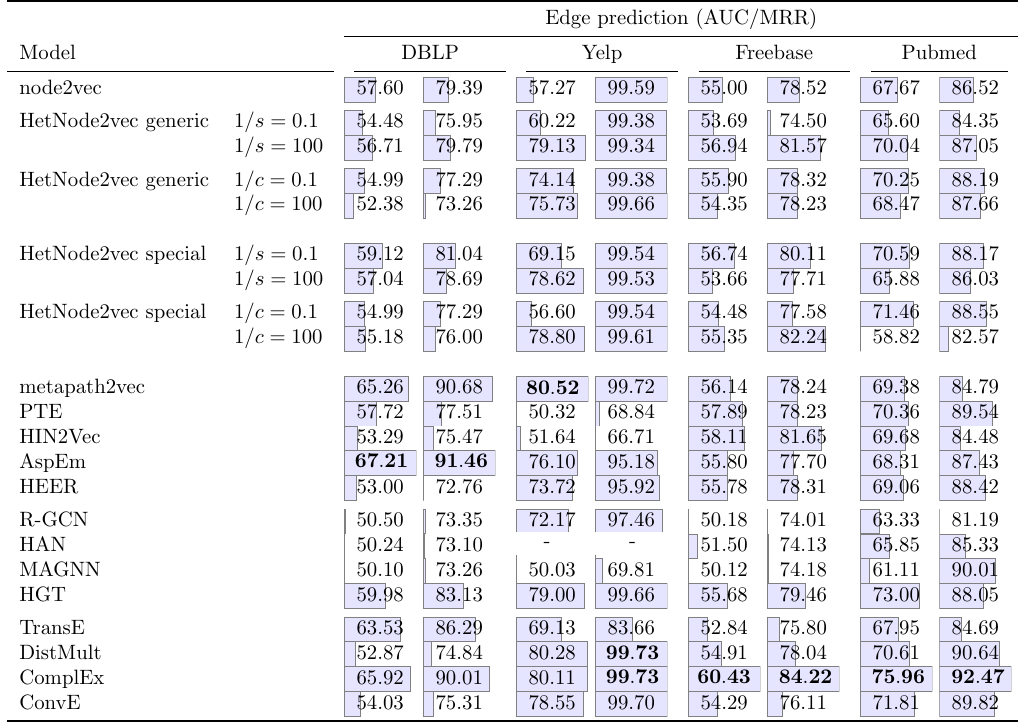}
\end{table}

%%%%%%%%%%%%%%%%%%%%%%%%%%%%%%%%%%%%%%%
\begin{table}[htb]
\caption{Performance metric for node label prediction and edge prediction tasks on the benchmark graphs using a Random Forest as the predictive model trained in the \Het graph representation.}
  \label{tab:random_forest_results}
      \centering
      \includegraphics[width=.9\textwidth]{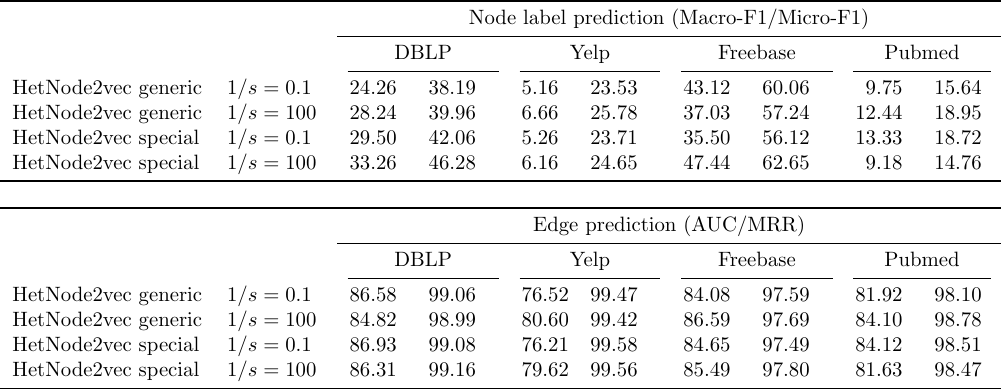}    
\end{table}

%%%%%%%%%%%%%%%%%%%%%%%%%%%%%%%

\section{Experimental hyperparameters}
\label{sec:hyperparameters}
\phantom{a}
 \begin{table}[h]
     \caption{Hyperparameters of the methods used in the construction of the embedding and their evaluation. While the default value for the number of RWs is 10 in Grape, we used five-fold cross-validation to assess whether a ten times larger value (no of RWs $= 100$) could improve the unsupervised embedding performance.}
    \label{tab:grape_skipgram}
    \centering
    \begin{tabular}{llr}
    \toprule
                                     & hyperparameter             & value                \\
      \midrule
      \multirow{4}{*}{RW generation} & Number of RWs (iterations) & 10-50               \\
                                     & $1/p$                      & 0.25                 \\
                                     & $1/q$                      & 4                    \\
                                     & Walk length                & 100                  \\
      \midrule
      \multirow{7}{*}{Skipgram}      & Embedding size             & 50                   \\
                                     & Epochs                     & 10                   \\
                                     & Ratio neg:pos samples      & 10:1                 \\
                                     & Window size                & 5                    \\
                                     & Max neighbours             & 100                  \\
                                     & Learning rate              & 0.01                 \\
                                     & Learning rate decay        & 0.9                  \\
      \midrule
      \multirow{4}{*}{Linear SVM}    & penalty                    & l2                   \\
                                     & tol                        & 0.0001               \\
                                     & C                          & 0.1, 0.001        \\
                                     & max iter                   & 3000                 \\ 
      \midrule
      \multirow{4}{*}{Random Forest} & Number of estimators       & 100                  \\
                                     & criterion                  & gini                 \\
                                     & min samples                & 1                    \\
                                     & max depth                  & None                 \\
                                     & max iter                   & 1000                 \\     
                                     & max features               & sqrt                 \\
    \bottomrule
    \end{tabular}
\end{table}

\end{document}